\documentclass[11pt,a4paper]{article}
\usepackage{helvet}

\usepackage[left=3cm, right=3cm, top=2cm, bottom = 2cm]{geometry}
\usepackage[utf8]{inputenc}
\usepackage[T1]{fontenc}
\usepackage[UKenglish]{babel}
\usepackage[UKenglish]{isodate}
\usepackage{lmodern}
\usepackage{paralist}
\usepackage{authblk}
\usepackage[table]{xcolor}

\usepackage{orcidlink}
\usepackage{tabularx}
\usepackage{booktabs}
\usepackage{doi}

\usepackage{rotating}
\usepackage{pdflscape}

\usepackage[colorinlistoftodos]{todonotes}

\definecolor{cambridgeblue}{rgb}{0.64, 0.76, 0.68}

\definecolor{cambridgeblue}{rgb}{0.64, 0.76, 0.68}

\begin{document}

\title{Engineering Reliable Autonomous Systems: Challenges and Solutions}

\makeatletter
\begin{titlepage}
   \begin{center}
       \vspace{2.5em}
       \begin{minipage}{0.45\textwidth}
       	\begin{flushleft}       
       		\includegraphics[width=0.8\textwidth]{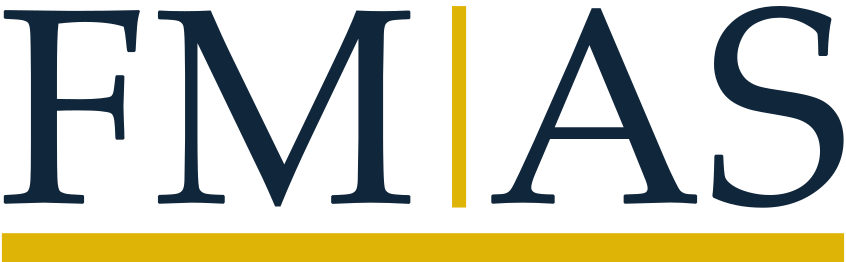}
       	\end{flushleft}
	   \end{minipage}
	   \noindent \begin{minipage}{0.45\textwidth}
	  	 \begin{flushright}
	   		\includegraphics[width=0.8\textwidth]{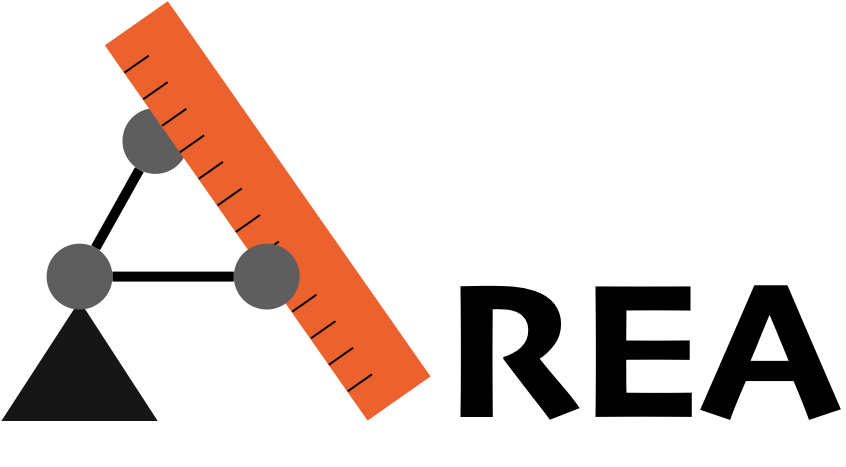}
	   	\end{flushright}
	   \end{minipage}

       \vspace*{5cm}

       {\LARGE \textbf{\@title}}\\
       ~\\
       \textit{WORKSHOP REPORT: \@date}

       \vspace{0.5cm}
       
            
       \vspace{1.5cm}

       \begin{center}
       \end{center}

       \vfill

   \end{center}
\end{titlepage}

\clearpage
\section*{Workshop Organisers}
\begin{itemize}
    \item Marie Farrell, University of Manchester, UK
    \item Matt Luckcuck, University of Nottingham, UK
    \item Angelo Ferrando, University of Modena and Reggio Emilia, Italy
    \item Rafael C Cardoso, University of Aberdeen, UK
\end{itemize}

\section*{Workshop Participants}
\begin{itemize}
    \item Natasha Alechina, Utrecht University, Netherlands
    \item Marco Autili, University of L'Aquila, Italy
    \item Diana Benjumea Hernandez, University of Manchester, UK
    \item Luciana Brasil Rebelo dos Santos, University of L'Aquila, Italy
    \item Daniela Briola, University of Milano-Bicocca, Italy
    \item Ana Cavalcanti, University of York, UK
    \item Christian Colombo, University of Malta, Malta
    \item Louise A. Dennis, University of Manchester, UK
    \item Clare Dixon, University of Manchester, UK
    \item Michael Fisher, University of Manchester, UK
    \item Mario Gleirscher, ENSTA | Institut Polytechnique de Paris, France
    \item Taylor Johnson, Vanderbilt University, USA
    \item Charles Lesire, ONERA, France
    \item Livia Lestingi, Politecnico di Milano, Italy
    \item Sven Linker, Kernkonzept GmbH, Germany
    \item Brian Logan, Utrecht University and University of Aberdeen, Netherlands and UK
    \item Colin Paterson, University of York, UK
    \item Fabio Papacchini, Lancaster University Leipzig, Germany
    \item Patrizio Pelliccione, Gran Sasso Science Institute, Italy
    \item Pedro Ribeiro, University of York, UK
    \item Maike Schwammberger, Karlsruhe Institute of Technology, Germany
    \item Silvia Lizeth Tapia Tarifa, University of Oslo, Norway
    \item Hazel Taylor, University of Manchester, UK
    \item Jim Woodcock, Southwest University, China; Aarhus University, Denmark; University of York, UK
    \item Mengwei Xu, Newcastle University, UK
    \item Yi Yang, KU Leuven, Belgium
    \item Huan Zhang, Maynooth University, Ireland
\end{itemize}
\clearpage

\section*{Executive Summary}

Engineering Reliable Autonomous Systems is an important new topic in Computer Science. As autonomous systems become more prevalent, ensuring there are easy-to-use techniques for building them reliably is increasingly important.

Two international workshop series that are leading the way in providing a venue for publication and discussion of state-of-the-art work in this field are the Workshop on Formal Methods for Autonomous Systems~(FMAS) and the Workshop on Agents and Robots for Reliable Engineered Autonomy~(AREA). Although related, these workshops have been situated in different sub-communities:~FMAS finds its home in the Formal Methods domain, whereas AREA garners attention in the Cognitive Robotics and Multi-Agent Systems community.

This workshop report captures and expands on the discussions in the Lorentz Center Workshop ``Engineering Reliable Autonomous Systems''~(ERAS) from the 10\textsuperscript{th} to the 14\textsuperscript{th} of June 2024\footnote{\url{https://www.lorentzcenter.nl/site/index.php?pntHandler=WorkshopTemplatePage&pntType=ConPagina&id=1982}}, which was co-organised by the organisers of both FMAS and AREA. The ERAS workshop brought together members of both the FMAS and AREA communities, industry practitioners, and representatives from sectors where the unique challenges posed by autonomous systems require the attention of our joint communities. This Lorentz Center Workshop unified the FMAS and AREA communities through their shared curiosity about the following research topics.
\begin{enumerate}
    \item Techniques for verification and validation of autonomous systems;
    \item Engineering real-world autonomous systems; and
    \item  Software architectures for safe autonomous systems.
\end{enumerate} 
One main outcome of the ERAS workshop is a catalogue of challenges in these areas and, most importantly, a pathway to solutions. 

Some challenges can already be tackled by techniques that are well known in academia, but have not had the required cut-through to become regularly used in practice. Other challenges remain untouched or unresolved -- despite already receiving attention from academia -- and require identifying the steps toward developing solutions. This roadmap is an important driver of future research work and industrial collaboration.


The workshop was held in person at the Oort venue of the Lorentz Centre, with a total of 30 participants over the five days. The workshop program included a mix of invited talks and breakout sessions. Some of our participants gave the invited talks to introduce and motivate the breakout sessions. The breakout sessions were based on the three research topics outlined above to identify the challenges and potential solutions in each topic. 

Tangible outcomes from the workshop included:
\begin{itemize}
    \item Cross-fertilisation of the FMAS and AREA communities via invited talks and detailed discussions in the breakout sessions;
    \item Co-creation of research roadmap and this workshop report;
    \item Follow-up events to be organised in other venues to maintain momentum; and
    \item A special theme in the journal \emph{Philosophical Transactions of the Royal Society A}, to include a summary of this workshop report, other papers resulting from ERAS workshop breakout sessions discussions, and invited papers that fit the topic of the theme.

\end{itemize}



\clearpage
\tableofcontents

\clearpage

\section{Introduction}

\textit{Engineering Reliable Autonomous Systems} is an important and rapidly emerging area within Computer Science. As autonomous systems become increasingly pervasive, the need for accessible and effective techniques for building them reliably is growing in parallel. Two international workshop series that play a leading role in showcasing state-of-the-art research in this area are the \textit{International Workshop on Formal Methods for Autonomous Systems (FMAS)\footnote{\url{https://fmasworkshop.github.io/}}} and the \textit{International Workshop on Agents and Robots for Reliable Engineered Autonomy (AREA)\footnote{\url{https://areaworkshop.github.io/}}}. While closely related in scope, these workshops are rooted in distinct research communities: FMAS is situated within the Formal Methods community, whereas AREA is primarily associated with the Cognitive Agent and Multi-Agent Systems community. The Lorentz Centre Workshop ``Engineering Reliable Autonomous Systems'' (ERAS) was held from the 10\textsuperscript{th} to the 14\textsuperscript{th} of June 2024, co-chaired by the organisers of FMAS and AREA. In this document we summarise the discussions from this workshop, outlining the current state-of-the-art, challenges and future directions for the field.

\subsection{Motivation}

Autonomous systems are becoming more prevalent as their physical capabilities (in the case of autonomous \textit{robotic} systems) and reasoning capabilities (including perception and cognitive) improve. 
Autonomy is already being used in areas like driverless cars \cite{5548121}, space exploration \cite{Cardoso2020}, nuclear decommissioning \cite{Bogue:11}, and domestic robotic assistants \cite{DWSFD15}. 
In some scenarios, autonomous systems offer benefits that classical automated systems cannot easily achieve. For example, in a hazardous environment, an autonomous system can reduce the need for human operators to enter unsafe regions, such as in the nuclear sector~\cite{Workington18}. In these mission- or safety-critical applications, the addition of autonomy complicates the prediction of the system's behaviour. 

As a result of the uncertainty arising from deploying systems with sophisticated reasoning capabilities in complex environments, the \emph{issue of autonomous systems assurance} remains unresolved. Much of the current state of the art mirrors approaches used in standard cyber-physical systems, but the move to true autonomy requires a step change in \emph{understanding and applying verification techniques} \cite{Luckcuck2019}. This broader development is needed for the field to advance and to safely harness the benefits of autonomous systems.

Most specific challenges in verifying autonomous systems can be categorised as either \textit{external} or \textit{internal} to the autonomous system under development \cite{Luckcuck2019}. The \textit{external} challenges include modelling an unpredictable physical environment and black-box components, and providing adequate evidence to gain public confidence and to demonstrate correctness and safety for certification purposes. 

The \textit{internal} challenges relate to the \emph{way that the system is developed}. Specifically, autonomous systems can generally be classified by their approach to AI:~symbolic (such as agent-based approaches to decision-making) and sub-symbolic (such as machine-learning techniques). This development choice essentially represents a compromise:~symbolic approaches are often more transparent and straightforward to verify, but sub-symbolic approaches are typically more scalable. Recent work on neuro-symbolic approaches attempts to bridge the gap between symbolic and sub-symbolic autonomy, but the \emph{challenges related to reliability} remain~\cite{sarker2021neuro}. Internal challenges arise in the use of (potentially heterogeneous) multi-robot systems that may also adapt or reconfigure themselves~\cite{Reconfig:Overview14}.

\subsection{The Workshop}

As a result of the assurance and verification challenges posed by autonomous systems, \textit{Engineering Reliable Autonomous Systems} has emerged as an important field of research in Computer Science. 
Many conferences have dedicated special tracks to topics related to autonomous systems\footnote{International Symposium on Formal Methods (FM), International Conference on Integrated Formal Methods (iFM), NASA Formal Methods Symposium (NFM), International Conference on Requirements Engineering (RE), etc. have all had special tracks related to autonomous systems in recent years.} and there has been a new conference series (since 2024) on exactly this topic, sponsored by the IEEE\footnote{ERAS Conference Series: \url{https://erasrobotics.org}}. Since the origins of our Lorentz Center workshop were from FMAS and AREA, we note that these two leading workshop series have significant overlapping topics. The ERAS Lorentz Center Seminar \emph{unifies the FMAS and AREA communities} through their shared curiosity of the following (combined) research topics:

\begin{enumerate}
    \item Techniques for verification and validation of autonomous systems,
    \item Real-world applications of autonomous systems, and
    \item Software architectures for safe autonomous systems.
\end{enumerate}

The ERAS workshop is an important first step in bringing the AREA and FMAS communities together and will help both communities individually. In particular, FMAS often lacks detailed \emph{industry-scale use cases} that typically appear at AREA. Conversely, AREA participants tend not to be fully aware of the benefits and state-of-the-art \emph{formal methods and techniques} for ensuring that autonomous systems behave correctly. This workshop report, produced collaboratively with participants at the ERAS workshop, provides the community with direction in this exciting and rapidly advancing domain. The ERAS workshop is the foundation for a \emph{joint research community}, built from these usually separate efforts, which will drive future \emph{networking grants} and \emph{project proposals}. Discussions prompted by the workshop and this workshop report will spawn \emph{new research collaborations}.

We recognise that many of the discussion points that are examined in this workshop report also apply to non-autonomous/conventional/traditional systems and some align with long standing systems engineering principles and practice. That said, many of these issues come to the fore with even stronger emphasis when systems are making decisions with little to no human oversight and so they are also emphasised in the context of autonomous system behaviour in this document.

\section{Case Studies}
\label{sec:caseStudies}
This section describes the 11 case studies that were used to drive the conversation at the ERAS workshop, and we present them here to contextualise and to aid the reader's understanding of the discussion summarised in Sections~\ref{sec:techniques}, \ref{sec:engineering}, and \ref{sec:architectures}.

\paragraph{Formal Modelling and Runtime Verification of Autonomous Grasping for Active Debris Removal.} This case study~\cite{farrell2021formal} concerns autonomous grasping for active debris removal in space, where a service vehicle equipped with a camera, robotic arm, and gripper must identify a suitable grasping point on a target object, capture it safely, and draw it closer without collision. The work is notable for combining architectural modelling, requirements formalisation, static verification, and runtime verification to analyse a pre-existing autonomous grasping system. Hardware and software components were modelled in AADL, requirements were elicited and formalised in FRET using a restricted natural language with temporal semantics, and verification was performed both statically with Dafny and at runtime with ROSMonitoring. The case study contains 20 requirements covering issues such as valid point cloud generation, optimal grasp point selection, collision avoidance, correct contact location, and force application. It presents results from both simulation and a physical testbed. These experiments showed that injected faults could be detected by runtime monitors and revealed gaps between requirements that appeared satisfactory in simulation and the limitations of physical hardware. More broadly, the case study illustrates both the value and the challenge of post-implementation verification for autonomous systems, showing that modularity and the interplay between implementation artefacts and verification artefacts are central to assuring real robotic systems in demanding domains.

\paragraph{ASUMI: Assuring the Safety of UAVs for Mine Inspection.} This case study~\cite{10.1007/978-3-031-72059-8_15} focuses on the assurance of unmanned aerial vehicles for infrastructure inspection in underground mines, where the aim is to remove humans from hazardous environments while ensuring that the autonomous system itself does not introduce unacceptable risks. The setting is especially demanding because mines are GPS-deprived environments where the UAV may experience failures such as signal loss, necessitating a safe autonomous response. The work centred on ensuring an autonomous return-to-home capability, enabling the UAV to retrace its path and land safely in the event of a failure. Key hazards included flying too close to surfaces, colliding with mine infrastructure or other objects, damaging the UAV itself, and, in the worst case, causing fire or explosion. To address this, the case study combined hazard analysis and risk assessment, derivation of safety requirements, dependent failure analysis, verification and validation activities, and confirmation of functional safety measures. A PX4 Vision Kit UAV was used in conjunction with laboratory experimentation and the Aloft simulation environment, which models mine settings, UAV control, and obstacles such as humans. Overall, the case study illustrates how safety assurance for autonomous systems in harsh real-world environments requires integrating realistic testbeds, explicit safety cases, and evidence that fail-safe autonomy can operate acceptably under constrained, safety-critical conditions.

\paragraph{DAISY: Diagnostic AI System for Robot-Assisted A\&E Triage.} This case study~\cite{10590219} concerns a proof-of-concept robot-assisted decision-support system for accident and emergency triage, developed in collaboration with clinicians and patients from York and Scarborough Teaching Hospitals NHS Foundation Trust to help address increasing pressure on admission times. DAISY is designed around a hybrid architecture in which a rules-based inference engine applies clinician-specified rules to process four categories of medically relevant information collected during triage: objective vital-sign measurements, demographic information, subjective symptom reports, and anatomical information. The system then provides clinicians with potential diagnoses, suggested investigations, low-level treatments, and referrals, while leaving the final judgement with the human clinician. A central feature of the case study is its emphasis on transparency and trust. Because the system is intended for use in a busy healthcare setting, the decision-making process was deliberately designed to be examinable and verifiable, so that its recommendations could be aligned with clinical reasoning and scrutinised by relevant stakeholders. More broadly, the case study illustrates the engineering challenges of deploying autonomous decision support in a safety-critical, human-centred environment, where explainability, clinician and patient trust, and the integration of autonomous reasoning into established professional workflows are all essential.

\paragraph{Agile Robotics for Industrial Automation Competition (ARIAC).} This case study~\cite{DBLP:conf/aiia/FerrandoKPCS020} uses the Agile Robotics for Industrial Automation Competition~\cite{10.1609/aimag.v39i4.2781}, organised by the US National Institute of Standards and Technology, as a realistic benchmark for studying the tension between agility, performance, and safety in industrial autonomous systems. The competition is set in a simulated warehouse built in Gazebo and deployed through ROS 2, where participants control a floor robot, a ceiling-mounted gantry robot, four automated guided vehicles, and a configurable set of sensors to perform tasks such as kitting, assembly, and pick-and-place operations. Within this setting, the case study focused on the runtime verification of a safety property concerning the separation distance between the gantry robot and a human operator, using ROSMonitoring to verify compliance with constraints derived from collaborative robotics standards. This makes the case study particularly useful as an example of how formal and runtime verification techniques can be embedded in realistic robotic workflows without removing the task's competitive and adaptive character. More broadly, it highlights the value of competitive environments as controlled yet nontrivial testbeds for engineering and assuring autonomous systems that must balance productivity, responsiveness, and human safety in industrial settings.

\paragraph{DARPA Assured Autonomy Case Study: Autonomous Underwater Vehicle (AUV).} This case study~\cite{muvva2021scitech} concerns an autonomous underwater vehicle for cable or pipe-following inspection missions, used as a representative example of assured autonomy in a highly challenging deployment setting. The system is based on a low-cost BlueROV platform. It operates in an environment whose dynamics are only partially known, with significant disturbances and sensing limitations, including ocean currents, sonar-based perception, communication constraints, GPS denial, and limited information update rates. To address these challenges, the case study adopts a high-level autonomy architecture that integrates sensing, perception, planning, control, actuation, environmental estimation, and world model creation. The primary mission objective is to follow a pipe or cable while avoiding obstacles, and this objective is decomposed using structured assurance artefacts such as bowtie diagrams and Goal Structuring Notation (GSN), alongside additional concerns including fault handling, battery management, plan management through behaviour trees, and runtime monitoring. Verification has been carried out for both the neural networks used for semantic segmentation of sonar data and aspects of the overall closed-loop system, complemented by simulation-based model validation and techniques for incorporating environmental uncertainty via online reachability. Overall, the case study illustrates the difficulties of assuring autonomous behaviour in a harsh and partially observable domain, while also showing how architectural decomposition, formal assurance arguments, and verification of learning-enabled components can be brought together to support real-world autonomous missions.

\paragraph{MASE: Modelling and Analysis in Mobility Software Engineering.} This case study~\cite{11334549} represents a broader research agenda on mobility software engineering that addresses several core challenges in autonomous driving through a combination of modelling, analysis, moral reasoning, and explainability. The work focuses on making the world machine-readable for autonomous vehicles by modelling spatial and temporal properties of traffic situations, detecting conflicts, and developing manoeuvre strategies, while also verifying desirable system properties such as safety, liveness, fairness, morality, and explainability. Within this agenda, abstract traffic models support logical reasoning about urban traffic scenarios, moral traffic agents aim to enable autonomous vehicles to resolve conflicts involving ambiguous rules, competing goals, and hazardous situations, and self-explainable digital twins seek to derive explanation models from collections of system models. Taken together, the case study shows that engineering real-world autonomous systems in mobility requires more than functional control alone. It also requires explicit attention to extra-functional concerns, such as moral decision-making and stakeholder-specific explanations. It is supported by modelling and verification approaches that capture the richness of traffic environments while remaining amenable to analysis.

\paragraph{Verification of a Domestic Robotic Assistant’s Behaviours.}
This case study involves the Care-O-bot, acting as a robot butler/assistant~\cite{DWSFD15,Webster16,GainerDDFHSW17,KWDGSFD21}. This is deployed in a domestic-type house (the robot house) at the  University of Hertfordshire. The robot house is equipped with sensors which provide information on the state of the house and its occupants. Low-level robot actions such as movement, speech, and light display are controlled by groups of high-level rules that together define particular behaviours. The robot is equipped with an interface that allows users to generate new behaviours using existing primitives. They focused on verifying properties of the system's decision-making using model-checking approaches. Additionally, they provided verification support for entering user-defined behaviours.

\paragraph{A Corroborative Approach to Verification and Validation of Human--Robot Teams.}
The case study concerns a collaborative manufacturing scenario in which the robot (BERT 2, Bristol Robotics Lab) and a person work together to assemble a table~\cite{WWADEMP19}.
The focus was on a table leg handover task from the robot to the human. The gaze, hand location and hand pressure of the human should be correct
before the handover takes place. A corroborative verification and validation approach was developed utilising different verification techniques (formal verification, simulation-based testing and physical robot experiments). Outputs from one technique were used to improve the models or requirements of the others.

\paragraph{Autonomous Fire-Fighting UAV.} This case study~\cite{DBLP:conf/mesa/ShaukatDKBMRARD24} concerns an autonomous unmanned aerial vehicle designed to search for a fire on the wall of a building, fly close to the detected fire, and spray suppressant liquid onto it. The system is based on a DJI M600 platform equipped with sensing and actuation components, including a RealSense D435i depth camera, an MLX90640 thermal camera, a two-axis nozzle gimbal, and a water pump. It uses a modular architecture that separates the main onboard computer from an Arduino that is responsible for pump and gimbal control. The engineering challenges are centred on accurate fire-source detection through vision-based perception, navigation and alignment with the fire centre, and safe extinguishing from an aerial platform. To support rigorous development, the system is modelled using RoboChart~\cite{10.1007/s10270-018-00710-z} and RoboSim~\cite{CSMRCDLT19}, enabling structured software and physical modelling, automatic generation of Gazebo simulation artefacts, and the prospect of code generation that preserves behavioural soundness. Verification and validation activities include search-based test generation from a digital model, conformance testing of observed UAV behaviour against expected behaviour, and verification of properties relating to the software, platform, and operational scenario. Overall, this case study illustrates the challenges of engineering and assuring a physically embodied autonomous system that must integrate perception, control, modelling, and verification in a safety-relevant real-world task.

\paragraph{Nuclear Inspection Case Study.}
This case study focuses on routine inspections within a nuclear facility, a critical scenario for maintaining the operational integrity of nuclear power plants. The operation involves conducting regular inspections in areas containing radioactive material to capture high-quality images for subsequent analysis and assessment, while ensuring safety and minimising human radiation exposure. After evaluating various options for this crucial task, the chosen solution is to operate the AgileX Scout mini robot autonomously~\cite{nuclearInspectionCaseStudy,AgileXRobot}. However, the use of AI in the nuclear industry is currently limited due to challenges in demonstrating its safety for specific operations. To address this, the proposed system architecture includes an intelligent Safety System operating in parallel with the control system~\cite{Louise&Chris2022}. These systems must be physically separate and independent, with sufficient redundancy and segregation to ensure reliability~\cite{SAPs}. Robust verification methods and rigorous testing are essential to guarantee the robot's predictable and safe behaviour~\cite{Farrell2018}.

\paragraph{The Human-(Multi)Robot Collaboration (CoHoMa) Challenge of the French Army.} The challenge is organised by ``BattleLab Terre'', a section of the French Army in charge of innovation and foresight, to study collaboration approaches between human operators and (semi-)autonomous systems on the field. There are human operators in a vehicle and robots on the field. The operators must reach successive territory lines without being detected. The scenario does not state which tasks must be performed autonomously. Some open questions remain regarding guarantees for implementing certain autonomous behaviours, such as stopping and warning the operator when navigation is impossible; autonomous backtracking when communication is lost; and co-disabling a trap with a robot and a drone~\cite{DBLP:journals/corr/abs-2310-07731}.

\paragraph{Summary.} Tables~\ref{tab:case-study-characteristics} and~\ref{tab:case-study-assurance} provide a comparative overview of the case studies discussed in this workshop report from two complementary perspectives. Table~\ref{tab:case-study-characteristics} focuses on the application setting of each case study, summarising the domain, the type of autonomy involved, the role of Machine Learning (ML), the place of humans in the loop, and the principal hazards or operational concerns. Table~\ref{tab:case-study-assurance}, in turn, focuses on how these systems are engineered and justified, comparing their regulatory or assurance contexts, the verification and validation techniques applied, the forms of evidence currently available, and the open problems that remain. Taken together, the two tables highlight both the diversity of real-world autonomous systems and the recurring assurance challenges that cut across domains, including uncertainty, human interaction, constrained operational assumptions, and the difficulty of connecting promising verification techniques to convincing evidence for deployment.

The workshop discussions suggested that case studies should play a more deliberate role in shaping future research. An ideal case study is not merely an example application, but an exemplar that helps the community evaluate validity, applicability, and generalisability. This requires greater clarity about which parts of a case study matter, which questions it can answer, and which remain beyond its scope. It also requires attention to the gaps between what current case studies allow us to analyse and what we would ultimately like to understand about real-world autonomy. Future work should therefore treat the design and selection of case studies as a research challenge in its own right, since progress in the field depends on having exemplars that are realistic enough to matter, yet structured enough to support rigorous comparison, analysis, and assurance.

\begin{landscape}
\begin{table}[p]
\centering
\scriptsize
\setlength{\tabcolsep}{4pt}
\renewcommand{\arraystretch}{1.15}
\begin{tabular}{p{3.2cm}p{2.0cm}p{3.0cm}p{2.3cm}p{3.2cm}p{4.2cm}}
\hline
\textbf{Case Study} & \textbf{Domain} & \textbf{Autonomy} & \textbf{Machine Learning} & \textbf{Human Role} & \textbf{Key Hazards / Concerns} \\
\hline

\rowcolor{lightgray!60} Autonomous Grasping for Active Debris Removal
& Space robotics
& Autonomous grasp planning and manipulation
& Not central
& Primarily engineering role
& Collision, failed grasp, incorrect contact point, unsafe force application \\

ASUMI
& UAV mine inspection
& Fail-safe autonomy with return-to-home
& Not central
& Human oversight remains important; humans removed from hazardous environment
& Collision, proximity to surfaces, loss of control, fire or explosion \\

\rowcolor{lightgray!60} DAISY
& Healthcare triage
& Decision-support autonomy
& Not central
& Clinician remains final decision-maker
& Misdiagnosis, inappropriate referral or treatment suggestion, loss of trust \\

ARIAC
& Industrial automation
& Task-level autonomy
& Not central
& Human operator represented in safety constraints
& Unsafe human-robot proximity, productivity versus safety trade-offs \\

\rowcolor{lightgray!60} DARPA AUV
& Underwater inspection
& Mission autonomy for pipe or cable following and obstacle avoidance
& Neural networks for sonar segmentation
& Human role is supervisory at mission-level
& Obstacle collision, loss of situational awareness, battery and fault-management failures \\

MASE
& Autonomous driving and mobility
& Driving decision-making and manoeuvre reasoning
& Not central
& Humans are road users and affected stakeholders
& Traffic conflict, unsafe manoeuvres, unfair or opaque decisions \\

\rowcolor{lightgray!60} Domestic Robotic Assistant
& Domestic robotics
& Rule-based behaviour selection and execution
& Not central
& End users can define behaviours; household occupants are direct stakeholders
& Unsafe or inappropriate behaviour in the home \\

Human--Robot Teams
& Collaborative manufacturing
& Shared autonomy in a handover task
& Not central
& Human directly in the loop
& Unsafe handover due to hand position or pressure mismatch \\

\rowcolor{lightgray!60} Fire-Fighting UAV
& Emergency response robotics
& Autonomous search in approach and extinguishing tasks
& Vision-based and thermal perception
& Human role is supervisory
& Misidentification of fire source, unsafe navigation, failed extinguishing \\

Nuclear Inspection
& Nuclear robotics
& Autonomous inspection for safety supervision
& ML use constrained by sector requirements
& Humans define mission goals and remain responsible for safety supervision
& Unsafe behaviour near radioactive areas, failed inspection \\

\rowcolor{lightgray!60} CoHoMa
& Defence and field robotics
& Human-multi-robot collaboration
& Not central
& Human operators coordinate mission-level objectives
& Navigation failure, communication loss, unsafe trap-disabling, mission compromise \\

\hline
\end{tabular}
\caption{Comparison of case studies by domain, autonomy characteristics, human involvement, and principal hazards or operational concerns.}
\label{tab:case-study-characteristics}
\end{table}
\end{landscape}

\begin{landscape}
\begin{table}[p]
\centering
\scriptsize
\setlength{\tabcolsep}{4pt}
\renewcommand{\arraystretch}{1.15}
\begin{tabular}{p{3.2cm}p{3cm}p{4.6cm}p{3.4cm}p{4.3cm}}
\hline
\textbf{Case Study} & \textbf{Regulatory / Assurance Context} & \textbf{V\&V Techniques} & \textbf{Assurance Evidence} & \textbf{Open Problems} \\
\hline

\rowcolor{lightgray!60} Autonomous Grasping for Active Debris Removal
& Mission assurance and safety-critical robotics
& Architecture modelling; requirements elicitation and formalisation; deductive verification; runtime verification; simulation and physical testing
& Formalised requirements, static proofs, runtime monitoring, and fault injection results
& Post-implementation assurance and transfer from simulation to hardware \\

ASUMI
& Functional safety and mining safety context
& Hazard analysis; risk assessment; safety requirements; dependent failure analysis; lab testing; simulation
& Safety-case-oriented evidence and realistic testing environments
& Stronger evidence for operation under harsh underground conditions \\

\rowcolor{lightgray!60} DAISY
& Clinical governance and medical regulation
& Rules-based reasoning; examinable and verifiable decision process; stakeholder review
& Transparency and alignment with clinical reasoning
& Wider stakeholder coverage, real-patient evaluation, and regulatory acceptance \\

ARIAC
& Human-robot collaboration and safety; and industrial robotics
& Runtime verification; simulation
& Runtime monitoring against safety constraints in realistic workflows
& Transfer from benchmark environments to industrial deployment \\

\rowcolor{lightgray!60} DARPA AUV
& Assured autonomy and safety-critical robotics
& GSN verification and validation; risk management; runtime verification; neural network verification; closed-loop analysis; simulation; online reachability tests
& Structured assurance artefacts and analysis of both ML and closed-loop behaviour
& Environmental uncertainty and partial observability \\

MASE
& Road traffic rules; moral and explainability concerns
& Abstract modelling; logical reasoning; formal analysis of safety, liveness, fairness, morality, and explainability
& Rich modelling of traffic situations and explainable digital twins
& Bridging abstract analysis and deployment-scale autonomy \\

\rowcolor{lightgray!60} Domestic Robotic Assistant
& Domestic assistive robotics and user safety
& Model checking of decision-making; verification support for user-defined behaviours
& Verification of behaviour rules and user-defined inputs
& Diverse stakeholders and changing home environments \\

Human--Robot Teams
& Human-robot collaboration and safety
& Simulation-based testing; physical experiments; corroborative V\&V
& Cross-checking and mutual refinement across methods
& Scaling corroborative V\&V beyond narrowly defined collaborative tasks \\

\rowcolor{lightgray!60} Fire-Fighting UAV
& Safety-relevant aerial operation, potentially in residential areas
& state machines; simulation; search-based test generation; conformance testing
& Model-based engineering and conformance testing
& Perception uncertainty, assurance of ML-enabled sensing, and operational approval \\

Nuclear Inspection
& Safe nuclear operations
& Program model checking; runtime verification; simulation
& Runtime monitoring, redundancy, segregation, and predictable behaviour
& Demonstrating acceptable assurance for autonomy in nuclear settings \\

\rowcolor{lightgray!60} CoHoMa
& Defence operational context
& Physical experiments; simulation
& Preliminary scenario-based evidence
& Allocation of autonomy and guarantees for key field behaviours \\

\hline
\end{tabular}
\caption{Comparison of case studies by assurance setting, verification and validation (V\&V) techniques, available evidence, and remaining open problems.}
\label{tab:case-study-assurance}
\end{table}
\end{landscape}

\section{Autonomous System Qualities and Desirable Capabilities}
\label{sec:challenges}

In this section, we discuss both the qualities that autonomous systems should have (Section~\ref{sec:qual})  and their desirable capabilities (Section~\ref{sec:capabilities}). In each section, we argue for specific qualities and capabilities that autonomous systems might need, in addition to those for classical systems.
Techniques for verification and validation of autonomous systems should take into account both qualities and capabilities (see Section~\ref{sec:techniques}). They also play a key role in the engineering of real-world autonomous systems~(see Section~\ref{sec:engineering}) and in identifying the main software architectures decisions for safe autonomous systems~(see Section~\ref{sec:architectures}).

\subsection{Autonomous System Qualities}
\label{sec:qual}

Often a system must be `high-quality' in some particular way (e.g. reliability or security) but in this section we argue that autonomous systems have specific quality dimensions that should be considered when specifying and evaluating them. The ISO/IEC 25010:2023~\cite{ISO/IEC25010} standard defines a model for software quality that identifies nine characteristics that software might have: functional suitability, performance efficiency, compatibility, interaction capability, reliability, security, maintainability, flexibility, and safety. 

In line with~\cite{Quatic2024}, modern systems
require a new understanding of quality and, consequently, new quality characteristics. 
Safety and security are both important quality characteristics; but \textit{in addition}, the following quality characteristics should be considered in the engineering of autonomous systems.
\begin{itemize}
\item \textit{Privacy}: data, especially personal data, should not be inappropriately available to others. It includes how data are collected, stored, modified, used, and exchanged between different parties~\cite{lu2023responsible}. In the case of AI-based systems, attention should also be given to training data.
\item \textit{Robustness}: the system should be able to continuing functioning (but possibly providing a lower level of service) if part of its hardware or software develops a fault, or if it receives incorrect input (either because of a sensor fault or because of a change in the environment).
\item \textit{Trustworthiness}: defined by four basic principles~\cite{EthicsGuidelines}: {\em respect for human autonomy}, {\em prevention of harm}, {\em fairness}, and {\em explicability}.

\item \textit{Sustainability}: the autonomous system should be engineered and built to not overly burden the environment with pollution or energy use. Further, they should benefit all human beings~\cite{EthicsGuidelines}, thus minimizing risk and avoiding negative impact for instance on democracy~\cite{Quatic2024}.	
\item \textit{Ethics/Responsibility/Accountability}: ensure that the system adheres to a range of ethical norms, constraints, and regulations~\cite{standard:bsi8611,EthicsGuidelines}.  
\item \textit{Social/Cultural Norms}: we should ensure that autonomous systems will abide by many of the social and cultural norms that humans do, especially when close human-robot interaction or human-robot teamwork is required.

\item \textit{Transparency}: closely linked to explainability (which we categorise as a desirable capability, see Section~\ref{sec:capabilities}) autonomous systems should be transparent in the sense defined in the EU AI Act~\cite{AI_Act_briefing}; to allow appropriate traceability and explainability, but also to make it clear when a human is interacting with an AI system and to not misrepresent the capabilities of the system. In general, universally accepted definitions of terms like explainability, transparency, interpretability, and understandability are yet to be found~\cite{10.1007/978-3-030-82017-6_8} so there is some overlap in usage in the literature.

\end{itemize}

\subsection{Autonomous System Desirable Capabilities}
\label{sec:capabilities}

To identify the desirable capabilities of autonomous systems, we refer to a recent work proposing an evaluation framework for autonomous systems, named LENS~\cite{TSEEvaluationFramework}. Through a deep analysis of the literature in autonomous and self-adaptive systems and robotics, LENS identified eight capabilities some of which are defined by sub-capabilities. 
The system is rated for each capability and subcapability so that an assessment using LENS shows what `level' the system is in each capability -- meaning to what extent the system exhibits this capability. 

The LENS framework identifies the following desirable capabilities.
\begin{itemize}
\item {\em Configurability} is the ability of the system to be configured to perform a task or reconfigured to perform different tasks. This may range from the ability to reprogram the system to alter its physical structure (e.g., by changing a tool).
\item {\em Adaptability} is the ability of the system to adapt itself to different work scenarios, environments, and conditions. Adaptation may take place over long or short time scales. Adaptability is divided into two sub-abilities: the {\em adaptation trigger}, which is what triggers the adaptation; and the {\em adaptation object}, which is the object being adapted and how the system alters its behaviour or structure (parameters, components, modes, or tasks) after adaptation.
\item {\em Dependability} is the ability of the system to perform its given task(s) without systematic errors. Dependability specifies the level of trust that can be placed on the system to perform.
\item {\em Autonomy} is the ability of the system to act autonomously, ranging from a simple autonomous task (e.g., when it reacts to sensor reading) to the ability to be self-sufficient in a complex environment.
\item {\em Interaction} is the ability of the system to interact physically, cognitively and socially either with users, operators or other systems around it. Interaction is composed of several sub-abilities: 
\begin{itemize}
\item {\em human-system interaction}, which concerns the interaction between users and the system;
\item {\em Human-system interaction feedback} to command a system depends on the user’s perception of the state of the system;
\item {\em system-to-system interaction} concerning the interaction between systems in carrying out a task or mission, and;
\item {\em human-system interaction safety} for those systems that have an inherent level of unsafety in the interaction between the human and the system.
\end{itemize}
\item {\em Perception} is the ability of the system to perceive its environment, by interpreting information and making informed and accurate deductions about the environment based on sensory data. Perception has two sub-abilities, {\em General perception} concerning the generic ability of a system to perceive environmental state by sensor data, and {\em Element recognition} ranging from being able to recognise instances of a single element, to being able to distinguish between many different elements or even identify elements that fit a generic pattern.

\item {\em Cognitive} is the ability of the system to interpret its task and the environment such that tasks can be executed even where the task or environment is uncertain. Cognitive contains several sub-abilities:
\begin{itemize}
    \item {\em Action} the ability of the system to act purposefully within its environment and the degree to which it can plan and carry out actions;
    \item  {\em Interpretive} the ability to interpret sensor data so that the system can identify, recognise, classify and parameterise elements in the environment; 
    \item {\em Envisioning} the ability to assess the impact of actions in the future; 
    \item {\em Acquired knowledge} the ability to reduce uncertainty about new situations by acquiring knowledge; 
    \item {\em Reasoning} the ability to reason from uncertain data and glue cognitive structures together, and; 
    \item {\em Cognitive human interaction} concerning the human interaction with a system that has a cognitive element.
\end{itemize}
\item {\em Explainability} concerning the ability of the system to make the entire control/adaptation process transparent and comprehensible by explicitly explaining it to humans. This is related to the quality of \textit{Transparency} (see Section~\ref{sec:qual}), which can be seen as an enabling quality for the capability of explainability. 

\end{itemize}

The qualities and capabilities discussed in this section provide a broader perspective on what it means for an autonomous system to be reliable and acceptable in practice. Traditional concerns such as safety, security, and dependability remain essential, but they are no longer sufficient on their own. Autonomous systems must also be trustworthy, transparent, robust, ethically aligned, and capable of operating effectively in complex environments while interacting with humans and other systems. Likewise, capabilities such as adaptation, perception, cognition, and explainability are not merely desirable features but increasingly important enablers of real-world deployment. Together, these qualities and capabilities offer a framework for understanding the expectations placed on modern autonomous systems and provide a foundation for the discussions that follow on verification and validation techniques, engineering practices, and software architectures for safe and reliable autonomy.

\section{Choosing Verification Techniques }
\label{sec:techniques}

This section discusses the problem of choosing a suitable verification technique for a given autonomous system. First, we give an overview of common verification techniques, and then discuss how to characterise the case studies described in Section~\ref{sec:caseStudies} in a way that makes selecting a verification technique easier. Finally, we conclude by discussing the challenges and directions in this area.


\subsection{Overview of Verification Techniques}
\label{sec:verificationTechniques}

There are several classes of verification techniques that one may choose. Section~\ref{sec:generalVT} gives an overview of the classes of techniques that have been using the case studies (Section~\ref{sec:caseStudies}) at either \emph{the system and component levels}. Then, Section~\ref{sec:NNVT} discusses techniques that are specifically used for verifying Neural Networks and other learnt systems. 

\subsubsection{General Verification Techniques}
\label{sec:generalVT}

Common to most verification efforts (largely independent of which technique is being used) is the need for a \emph{specification} that reflects the requirements (i.e. the notion of correctness or performance) for each of the considered characteristics.  A fundamental pattern that many specifications follow is what is usually known as the \emph{assumption/guarantee} style: the system \textit{assumes} that some properties hold, and in turn will \textit{guarantee} that after execution some other properties hold. 

At design-time, \emph{model checking} and \emph{theorem proving} allow one to verify that a system obeys given properties using models or theories; that is, property-specific abstractions of certain system characteristics e.g. changing decision-making components, uncertainties in human-system-interaction. For example, probabilistic model checking can deal with uncertainties as `first-class citizens' in such abstractions.  Hybrid systems theorem proving enables the precise reasoning about low-level control aspects of an autonomous system.  Certain forms of change can, for example, be tackled by incremental model checking.  

Moreover, \emph{static analysis} (e.g. assertion-based program verification) can be used to directly verify the system's software components.  \emph{Correct-by-construction synthesis} supports the algorithmic derivation of, for example, decision-making components of an autonomous system.  Furthermore, \emph{verification} (e.g. certain forms of model checking) can be performed \emph{at run-time} in cases where no adequate models of the system are available at design time.  Additionally, an implementation can be validated by various methods of \emph{testing} (e.g. in a HIL simulator or even by statistical model checking).

\subsubsection{Verifying Neural Networks}
\label{sec:NNVT}

There are many approaches and significant research that aims to verify that a trained neural network satisfies a specification, known as neural network verification, for example~\cite{tran2022mdat,seshia2022cacm}. 
To drive innovation in this topic, the Verification of Neural Networks Competition (VNN-COMP)\footnote{VNN-COMP: \url{https://vnn-comp.github.io/}} collects benchmarks and evaluates many of these recent approaches, with recent summary overviews of software tools implementing these techniques~\cite{brix2023sttt}.
These vary from classical formal methods style approaches such as satisfiability and Satisfiability Modulo Theories (SMT) as well as abstract interpretation, to constrained optimization approaches such as with Mixed Integer Linear Programming (MILP) and other solvers, along with control theoretic techniques like reachability~\cite{liu2021now,albarghouthi2021now}.

For control systems that use neural networks as feedback controllers where a specification is defined over the states of a plant model, known variously as intelligent control systems, Neural Network Control systems (NNCS), neural feedback loops, among others, the ARCH-COMP AI and NNCS category report surveys many of the recent techniques and advances for these classes of learning systems~\cite{lopez2023archcomp_ainncs}. 
Additionally, there are verification approaches for different classes of machine learning models as well, such as decision trees, Markovian models, etc.

Rather than verifying the results of learning, some researchers attempt to constrain the results of learning in advance. For example, \cite{GiunchigliaSL22} contains a review of work on \emph{deep learning with logical constraints} (the results of learning are probabilistically guaranteed to respect given logical constraints). Similar work exists in reinforcement learning, for example \emph{shields} proposed in \cite{Alshiekh//:18a} provably constrain the behaviour of a Re-enforcement Learning (RL) agent during training and deployment to satisfy safety constraints expressed in Safety Linear-Time Temporal Logic. Current challenges are scalability, limited expressivity of properties in case of neural networks and production of correct and useful abstractions of the learning environment in case of RL.

\subsection{Case Study Characteristics} 
\label{sec:caseStudyChars}

Here, one sub-group of the workshop contributors describe an approach that makes the case studies easier to compare, by defining several criteria and properties to characterise the system. The group rated each case study with a score between 0 and 5 for each of the criteria and properties. Note: this is the sub-group's subjective assessment but it provides a useful illustration of how to compare systems and select verification techniques.


We defined the following criteria for comparing the case studies:
\begin{itemize}
    \item \textit{Openness}: how controlled is the environment in which the case study operates?
    \item \textit{Risk}: severity of a potential failure multiplied by the probability of the failure happening.
    \item \textit{Change}: how frequently and significant are changes in the environment, the requirements, and the implementation expected?
    \item \textit{Software Complexity}: how complex is the software architecture of the case study?
    \item \textit{Hardware Complexity}: how complex is the hardware architecture of the case study?
    \item \textit{Perception Complexity}: how complex is the perception component of the case study?
    \item \textit{Regulation}: how much/how stringent the regulatory environment is for the sector in which the system will be used.
    \item \textit{Uncertainty}: how sure are we about the system's environment?
    \item \textit{Human Control}: what level of human supervision and control can be exerted over the system?
    \item \textit{Human Take-Over Time}: how quickly would a human have to take over control from the system, if requested?
\end{itemize}

These characteristics consider both the system itself and assumptions about the environment (relating to both the internal and external challenges discussed in Luckcuk \& Farrell et al.~\cite{Luckcuck2019}). Both types of characteristic have repercussions for verifying the system.

We also assess the case studies for the \textit{properties} that they exhibit, here properties are more general types of system requirements such as safety or security.
Autonomous systems have the potential to be used in various environments, for example underwater, in space, on extraterrestrial planets or moons; and in various sectors, for example firefighting, cleaning and rubbish collection, manufacturing, the nuclear industry, or autonomous vehicles.
Despite the diversity of domains and environments, these autonomous systems are still likely to have to obey properties like safety or security, just like non-autonomous systems.

We asses to what extent each case study might need to be checked for the following properties.
\begin{itemize}
    \item \textit{Safety}: something bad never happens. 
    \begin{itemize}
        \item \textit{Passive safety}: the system remains safe if it does not perform any actions. 
        \item \textit{Active safety}: the system only performs actions that do not intentionally lead to an unsafe state. 
    \end{itemize}
    \item \textit{Security}: the system is robust against cyber attacks. 
    \item \textit{Liveness}: something good will eventually happen. 
    \item \textit{Fairness}: ensures each agent of the system will get the chance to get resources. 
    \begin{itemize}
        \item \textit{Neural Network Output Fairness}: it is important to user a fair, unbiased data set when training a neural network.
        \item \textit{Task Fairness}: each task gets fair access the the processor and critical resources, avoiding task starvation. 
    \end{itemize}
    \item \textit{Social, Legal, Ethical, Empathetic, and Cultural (SLEEC) properties}: a broad range of properties defined by Townsend et al.~\cite{DBLP:journals/mima/TownsendPANCCHT22} that define how an autonomous system should behave to obey social, ethical or cultural norms; to obey legal requirements; or to act empathetically. These are especially important as autonomous systems are being proposed for using in people's homes, or other personal settings.
\end{itemize}

\begin{figure}[t]
\begin{center}
\scalebox{.5}{
\includegraphics{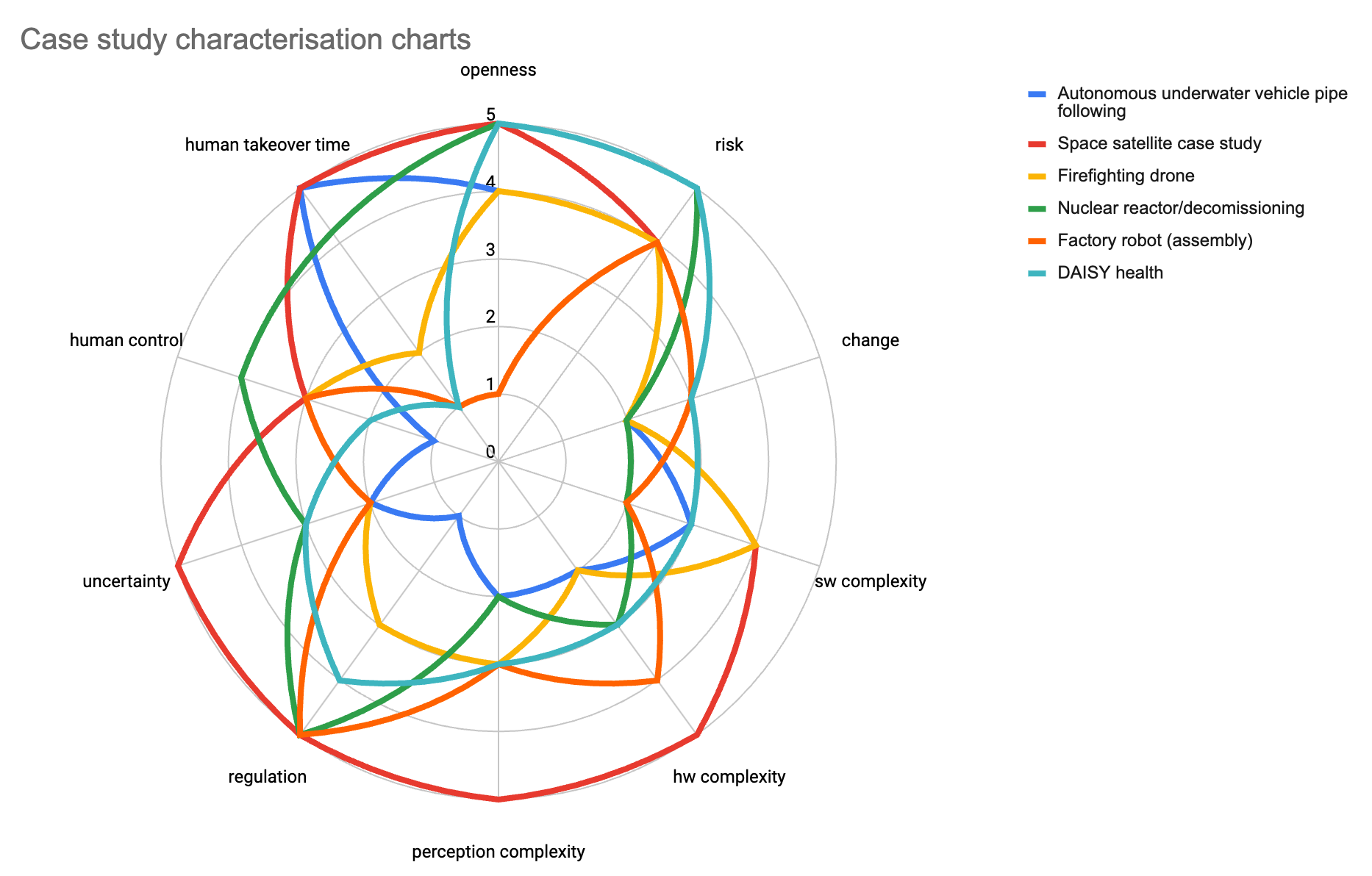}
}
\end{center}
\caption{Radar plot of our subjective assessment of the case studies in Section~\ref{sec:caseStudies} against our 10 characteristics. Note: the numbers represent discrete ratings rather than a numerical scale.} \label{fig:radars}
\end{figure}

The radar plot in Figure~\ref{fig:radars} visualises our subjective assessment of the case studies from Section~\ref{sec:caseStudies}, which highlights the similarities and differences between the case studies. 
All the case studies show moderate frequent/significant change in their environment (rated 2 or 3), and high levels of potential risk~(rated 4 or 5). However, there are large differences in the required human take-over time~(with some case studies rated 5 and some rated 1 or 2) and the level of regulation in the sector~(the case studies were rated 5, 4, 3, or 1). 

\begin{figure}[t]
\begin{center}
\scalebox{.5}{
\includegraphics{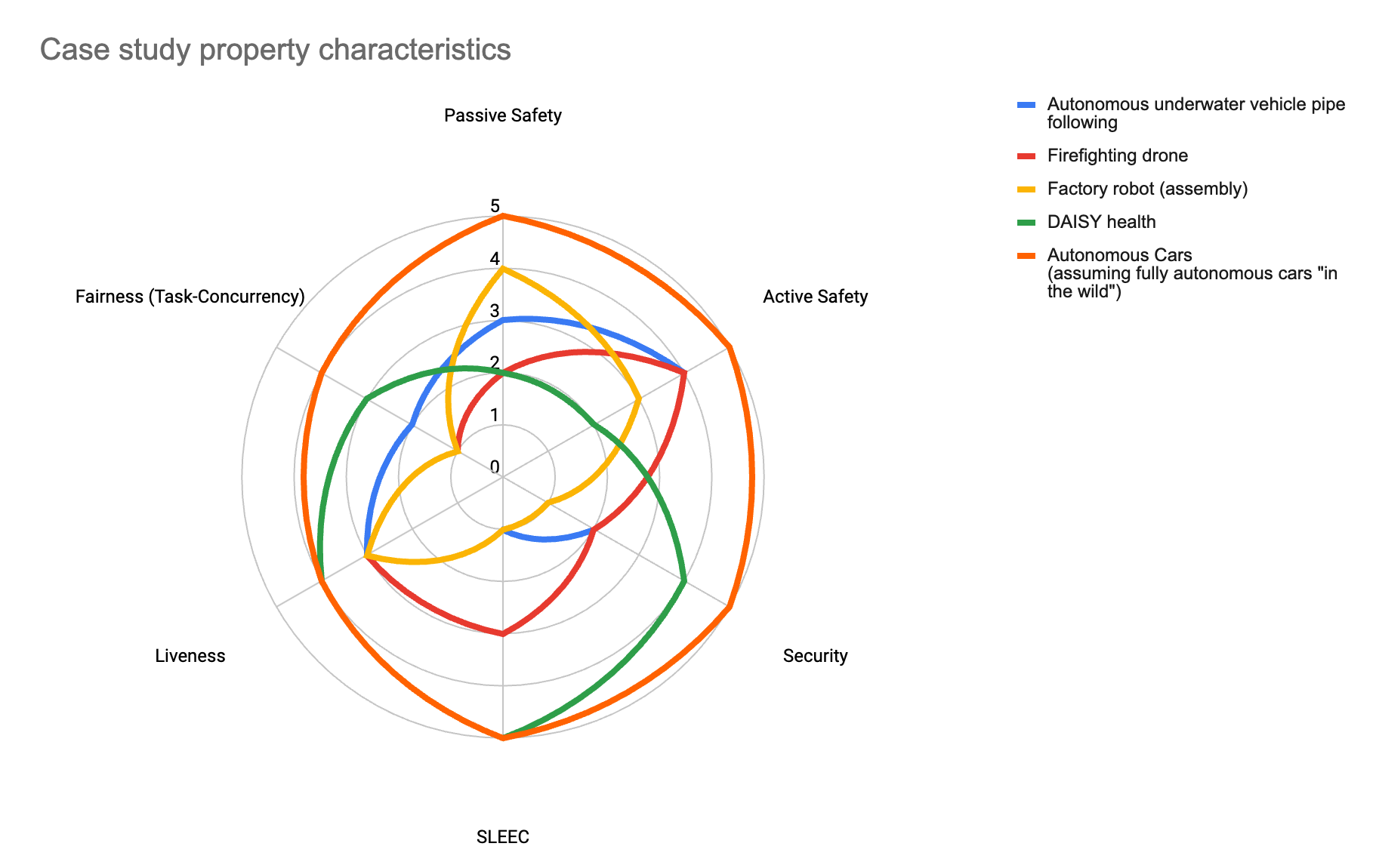}
}
\end{center}
\caption{Radar plot of our subjective assessment of the case studies in Section~\ref{sec:caseStudies} against our six properties. Note: the numbers represent discrete ratings rather than a numerical scale.
} \label{fig:radarprops}
\end{figure}

Figure~\ref{fig:radarprops} shows the radar plot of our subjective assessment of the case studies against the list of properties. We can see that (fully) autonomous cars exhibit high need for all of the properties, because they would be operating in a relatively open environment around humans. However, the firefighting drone and factory robot show much smaller areas within the plot because of their constrained environments and limited actions.

The examples in Figures~\ref{fig:radars}~and~\ref{fig:radarprops} show how this approach can be used to compare different systems, using our case studies. 
However, the radar plots can also be useful when selecting a verification technique, provided that radar plots are produced for the verification techniques themselves.

Different verification and validation techniques are more or less suitable for handling particular characteristics. Comparing the shape of the radar plots for a system and a proposed verification technique can 
give a quick indication of whether the technique is appropriate for the system. 
For example runtime verification can be used as a security wrapper around a high-risk system to suppress dangerous manoeuvres; while, bounded model checking might be a natural choice for the verification of complex perception. If these techniques are assessed against how capable they are at handling each characteristic (both criteria and properties), then their suitability for the system being built can be quickly ascertained.


\subsection{Future Challenges and Directions}

\paragraph{Change}
Autonomous systems have to contend with change throughout their life-cycle, such as, operating in environments or scenarios that differ from those originally envisioned. These situations can invalidate results of V\&V, for example, by operating outside known operational assumptions, and also limit the extent to which V\&V effort can be reused to meet new operational requirements. Moreover, properties of interest may to be revisited in response to: experience obtained through deployment; changes in the regulatory and legal landscape; and evolving cultural and societal views on the use of autonomous system.

To address change, architectures that enable leveraging of integrated, compositional, and incremental V\&V techniques, should be explored, potentially specialised by application domain, given the range of characteristics considered for autonomous systems. This will help to minimise the effort in re-establishing conformance/assurance.



\paragraph{Uncertainty and Learning}

Uncertainty in autonomous systems arises when its behaviour diverges from anticipated outcomes due to the dynamic and unpredictable nature of various factors inherent in software systems, whether these factors are related to the system itself, or to sensors inputs, environment distortion, erroneous human behaviour, to name a few. Uncertainty often becomes a problem when the system has to make a decision because the uncertainty might case a failure in the specified behaviour.

To address uncertainty, the autonomous system must possess the capability to adjust its behaviour, preferably in real-time, whenever relevant knowledge becomes available. As stated in \cite{mahdavi2017classification}, a viable software-based solution to the problem of uncertainty lies in equipping the system with self-adaptation capabilities. These adjustments often need to maintain the original level of performance, which may be difficult. To summarise, when designing autonomous systems, the intertwined relation to the environment, humans-in-the-loop and the complex interaction of the heterogeneous models should to be taken into account to deal with this high level of uncertainty as a critical factor.

Systems that learn the behaviour to perform also present challenges for verification; and although there is promising work in this area, there are many open research problems.  

Learning correct (satisfying safety and other specifications) systems also has multiple challenges, ranging from scalability and difficulty of producing useful abstractions to even defining properties of interest (e.g. `avoid collisions with pedestrians' or `be polite and let people pass when culturally appropriate').

\paragraph{Combining Verification Techniques and Results}
Currently there are many different verification techniques, both formal and non-formal, used for the assurance of systems. We believe looking at how these can be used {\em together}  is a key area for tackling autonomous systems verification. We see two ways to apply different verification techniques to the same system: 
\begin{itemize}
    \item applying a combination of two (or more) techniques to the same system or subsystem; or,
    \item applying different verification techniques, in concert, to different subsystems.    
\end{itemize}

The former approach uses several verification techniques on the same system/sub-system, which prompts the question: `how do we combine outputs from different types of technique together to improve the overall confidence in the system?' For example can an error identified from the formal verification (model checking, theorem proving etc.) be explored further in testing or end user experiments (see for example~\cite{WWADEMP19}). Autonomous robotic systems pose a variety of challenges that need a combination or integration of several formal methods~\cite{Farrell2018}.

\begin{figure}[t]
    \centering
    \includegraphics[scale=1]{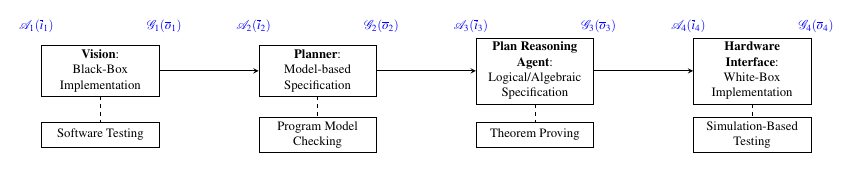}
    \caption{Illustration, from \cite{farrell2019modular}, of an approach to link system components (or subsystems) using assumption/guarantee specifications, while enabling each component to be verified by the most suitable technique.}
    \label{fig:agreasoning}
    \end{figure}

The latter approach uses different verification techniques for different subsystems, and so must try to coordinate these different techniques at the macro level. 
For example, the work on modular verification~\cite{farrell2019modular, Cardoso2020} where each subsystem has a macro-level assumption/guarantee specification in first-order logic, but can be verified using different techniques. Figure~\ref{fig:agreasoning} (from \cite{farrell2019modular}) illustrates this, showing four components (subsystems) of an autonomous rover and their associated micro-level verification techniques; each component is linked by its macro-level assumptions and guarantees.

In general, how to coherently bring together the outputs of different verification techniques remains an open question. For example, do the results from different techniques provide different levels of confidence in those results? If so, can we quantify some overall confidence? Also, how do the results from various formal and non-formal verification techniques feed into a single, defensible assurance argument. Are we sure that the application of different techniques to different subsystems, or applying a combination of techniques to the same subsystem, are verifying the same property?

\paragraph{Specification Framework for Autonomous Systems}

We see the need for an overarching specification style allowing autonomous systems assurance to be structured in a more systematic way.  For example, the aforementioned assumption/guarantee style can serve as a basic template for this.  Variants such as temporal/timed assumptions and timed probabilistic guarantees can serve as a means to break down the requirements for each of the considered characteristics. Specification languages should also be developed to describe elements of autonomous systems that are not captured by existing approaches, such as how specify perception components and an overall computer vision pipeline's correctness.
This may involve concepts like scene graphs, knowledge graphs, along with broader classes of logic, such spatial logics, probabilistic logics, dense-time temporal logics, and beyond, so that we can characterise formally behaviours of systems even when we know they will not always be correct, but may operate reliably with some probability, at least on datasets with similar distribution to those that were used for learning.
How to compose perception with other system components necessitates an assumption/guarantee approach, when such a perception pipeline is used along with sensor fusion, planning, and control.

\section{Engineering Real-World Autonomous Systems}
\label{sec:engineering}

\begin{figure}
\centering
\includegraphics[width=0.8\linewidth]{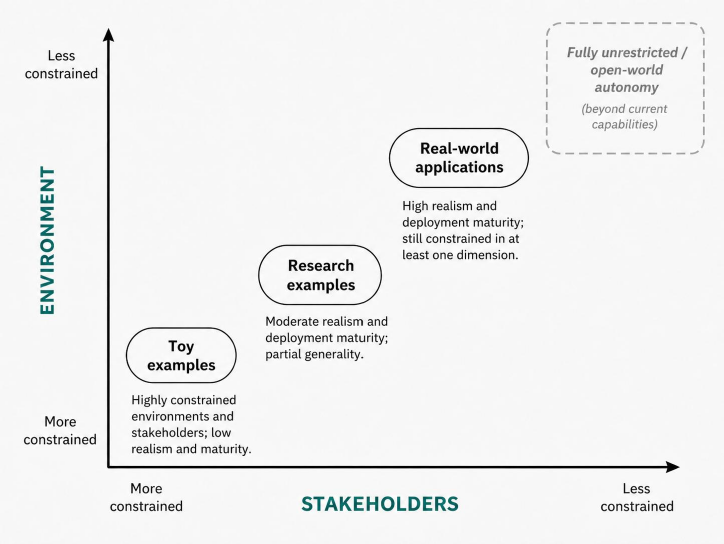}
    \caption{Most deployed autonomous systems operate under significant environmental, stakeholder, or combined constraints, rather than in fully open real-world settings.}
    \label{fig:reality}
\end{figure}

Engineering real-world autonomous systems remains challenging precisely because many systems described as ``real-world'' are still deployed only under substantial constraints. In practice, autonomy is rarely exercised in a fully open setting. Instead, most systems operate within bounded environments, with restricted stakeholder interactions, simplified functionality, or carefully controlled operational assumptions, as illustrated in Figure~\ref{fig:reality}. This can also be viewed through the lens of an Operational Design Domain (ODD), in which a system is expected to operate reliably only within a specified set of conditions, assumptions, and constraints. Many deployed autonomous systems are therefore reliable not in an unrestricted sense, but only within a carefully delimited operational envelope. These constraints are not necessarily shortcomings. In many cases, they are deliberate and appropriate engineering choices that make deployment feasible, safe, and certifiable. However, they also highlight an important gap between demonstrating autonomy in a realistic setting and engineering autonomous systems that can robustly cope with the full complexity of the real world.

The case studies introduced in Section~\ref{sec:caseStudies} help to make these distinctions concrete. DAISY, for example, engages with genuine clinical practice through inputs from real clinicians and real sensing modalities, but it excludes important stakeholder groups such as patients, technicians, and other clinicians, and has not yet been evaluated with real patients. Similarly, the fire-fighting drone involves a physical platform and real operational functionality, yet it remains far from fully unconstrained deployment due to simplified settings, unresolved perceptual uncertainty, and the need to meet demanding engineering and regulatory standards. Other case studies reveal different kinds of restrictions. The solar farm maintenance robot is motivated by a real deployment context and existing hardware, but the operational environment is fenced, and human interaction is not a primary focus. Household robotic assistants, by contrast, are strongly shaped by the complexity of their environments and the diversity of stakeholders they affect, including adults, children, and pets, which raises challenges for trust, empathy, adaptation, and explainability.

These examples show that the engineering challenge is not simply to move from simulation to reality, but to understand which aspects of reality are already captured, which remain abstracted away, and what that means for assurance. Real-world autonomous systems may be constrained by stakeholders, the environment, or the range of functionality they are expected to deliver. Each of these forms of restriction has consequences for how evidence is gathered, how safety and trustworthiness are argued, and how responsibility and liability are understood. For this reason, the discussion in this section connects directly to the broader challenges identified in Section~\ref{sec:challenges}, while focusing specifically on those that emerge when autonomy must be engineered for real-world deployment.

Our discussion is organised around three complementary perspectives: Practical Case Studies, Scientific Advancements, and Assurance. Across these perspectives, a recurring theme is the need to identify safety- or mission-critical scenarios in which high-level autonomous decision-making must be engineered with particular rigour. This is especially important when decisions depend on complex machine learning components, rich domain knowledge, and the often competing requirements of multiple stakeholders. In such settings, engineering discipline is essential not only for system performance but also for understanding liability, managing potential harms, and making the impact of autonomous behaviour sufficiently transparent and defensible.

\subsection{Challenges of Engineering Real-World Autonomous Systems}

\paragraph{Explainability}
Explainability is a central requirement for real-world autonomous systems because these systems interact closely with a wide range of stakeholders, including end-users, domain experts, technicians, regulators, and, in some cases, lawyers or investigators. In this context, explanations serve several important purposes. They can increase transparency for regulators not only with respect to the system’s behaviour but also with respect to the engineering process itself, for example, by clarifying whether sufficient testing was carried out or whether the chosen abstractions were appropriate. Explanations can also help end-users to develop a better understanding of the system, which is especially important in shared-task settings where human and autonomous actions must be coordinated to preserve safety. Beyond this, explainability can support trust and user acceptance by helping people to understand why the system acted as it did. It also enables traceability of behaviour and faults, which is particularly relevant in the context of tort claims, post-incident investigation, and questions of liability.

\paragraph{Modelling Language Expressiveness}
Real-world autonomous systems exhibit substantial complexity, with rich structural representations and sophisticated semantic behaviours that may involve data modelling, temporal dynamics, concurrency, hybrid discrete-continuous dynamics, and environmental uncertainty. Formal modelling and analysis are widely recognised as an effective means of supporting rigorous engineering and providing evidence for safety assurance. Although existing formal modelling and analysis languages have achieved notable success, particularly in hardware and more conventional software systems, they were not primarily designed for autonomous systems and therefore remain limited in their expressiveness in this setting. In particular, they often struggle to capture the full range of behaviours required in autonomous systems, as well as properties that relate not only to functionality and safety but also to social, legal, and ethical considerations. There is therefore an urgent need for modelling languages for autonomous systems that are expressive enough to represent this broad range of relevant aspects while remaining computationally tractable to support rigorous analysis.

\paragraph{Requirements Elicitation and Testing}

Requirements elicitation and testing are especially critical when engineering distributed autonomous systems for real-world deployment, because many of the most important challenges must be addressed at the requirements stage rather than deferred to later development phases. In such systems, the range of stakeholders is often unusually broad, extending beyond traditional clients and operators to include humans affected by the system, whether directly involved or not, as well as legal experts, regulators, and standards bodies. At the same time, these systems may rely on multiple machine learning components whose capabilities and limitations are not always well understood by non-experts, making it important to involve appropriate technical expertise early to ensure that requirements remain realistic and feasible. Testing also becomes more demanding, as acceptance tests must cover not only functional and non-functional requirements but also realistic deployment scenarios that reflect the complexity of the operational environment. However, this is complicated by the fact that complete ground truth may be unavailable in partially unpredictable real-world settings, meaning that uncertainty and resilience must themselves be treated as core requirements. In addition, testability requirements should be specified explicitly so that the resulting system supports meaningful explainability, diagnosis, and assurance. Finally, real-world autonomous systems are typically subject to a wide range of non-functional requirements, many of which are difficult to verify formally, so particular care is needed to formulate them in ways that can later be translated into analysable models and verification artefacts.
  
\paragraph{Requirements Beyond Functionality and Safety}
Engineering real-world autonomous systems requires a broader view of requirements than is typical in more conventional software systems. In addition to core functional behaviour and safety properties, these systems must satisfy requirements connected to the contexts in which they operate, the people and organisations they affect, and the standards and norms that govern their deployment. This means that requirements engineering must account not only for what the system should do, but also for whether it is appropriate, acceptable, understandable, and justifiable in its intended setting. Relevant stakeholders may include end users, affected communities, regulators, legal authorities, and domain experts, each of whom may impose different expectations and constraints. As a result, the specification, validation, and verification of real-world autonomous systems must address a wider spectrum of properties than traditional correctness alone. One useful way of structuring this broader space is through SLEEC properties, which capture social, legal, ethical, empathetic, and cultural requirements. These provide a lens for addressing the full range of concerns that may need to be reflected in autonomous system behaviour, as explored in recent work on translating pluralistic normative principles into autonomous agent rules~\cite{DBLP:journals/mima/TownsendPANCCHT22} and on the specification, validation, and verification of SLEEC requirements for autonomous agents~\cite{DBLP:journals/jss/YamanRCCPT25}.

\paragraph{Uncertainty}
Uncertainty is a central challenge in engineering real-world autonomous systems because the forms of uncertainty that arise in perception are not usually captured well by traditional verification techniques. To verify such systems in a way that preserves the intent of their goals, we may need new methods for representing and reasoning about uncertainty in perception and interpretation. This includes familiar problems such as misclassification, inaccurate bounding boxes, or fragmented regions in medical scans, but the impact of these errors is highly context-dependent. For example, misclassifying a give-way sign as a stop sign may be far less serious than misclassifying it as a 30 mph sign, which suggests that uncertainty should not be treated as a purely statistical quantity detached from operational consequences. A further difficulty is that performance measures are typically derived only from the data that is used for model development, testing, and verification, with no guarantee that these data are truly representative of the world in which the system will operate. In practice, autonomous systems often must reason over imperfect data for which the ground truth may be unknown or only partially available. This problem is compounded by out-of-distribution phenomena, including shifts in the input space caused by real-world perturbations or underrepresented situations (e.g., people carrying umbrellas) and shifts in the label space where entirely new object categories or environmental features appear over time (e.g., new forms of road furniture). More broadly, autonomous decision-making often depends not only on object recognition, but also on inferring behaviour from uncertain symbolic representations. For instance, a system may initially identify a person depicted on the side of a moving bus and incorrectly infer that a pedestrian is crossing the road. Capturing and analysing this kind of uncertainty remains an important open challenge for the rigorous engineering and assurance of real-world autonomous systems.

\paragraph{Interaction}

Interaction is a fundamental concern in the engineering of real-world autonomous systems because such systems do not operate in isolation but exist within environments they must perceive and respond to, often interacting with humans or other autonomous systems. For this reason, the nature of these interactions should be clearly defined, understood, and agreed upon during requirements elicitation. This includes specifying which aspects of the environment must be perceived, at what level of abstraction, and with what degree of reliability, as well as characterising interactions with other entities, which may be cooperative, supervisory, competitive, or adversarial. These issues are reflected across the case studies discussed in this section. In DAISY, for example, interaction is shaped by clinical workflows, sensor inputs, and medical professionals' roles in interpreting and acting on system outputs. In the fire-fighting drone case study, interaction extends to a dynamic physical environment and to operational constraints that may involve coordination with human responders and safe behaviour in residential settings. More broadly, interaction design has direct implications for safety, trust, responsibility, and assurance, since poorly specified interactions can lead to mismatched expectations, unsafe behaviour, or failures that are difficult to interpret and attribute.

\paragraph{Relations of Space and Time}

Relations of space and time are fundamental to the engineering of real-world autonomous systems because they shape what the system is concerned with, how requirements should be formulated, and which forms of reasoning and action are feasible. Depending on the application, an autonomous system may need to reason primarily about space, primarily about time, or about both, either as largely separate dimensions or as tightly entangled aspects of the task. These distinctions matter because notions of time and space are central to perception, reasoning, and action planning. They influence the level of perception required, the abstractions acceptable, and the degree of reasoning complexity that can be tolerated before it begins to undermine real-world effectiveness. In practice, there is often a trade-off between richer spatio-temporal representations and the need for timely decision-making, especially when approximations may affect physical actions in the environment. The relevance of these issues varies across case studies. In DAISY, for example, space is not a dominant concern, and the system operates under relatively relaxed timing constraints. By contrast, in the fire-fighting drone case study, both spatial and temporal considerations are central, since the system must reason about its distance from the fire and how quickly it can reach the target. More broadly, the treatment of space and time has direct implications for requirements engineering, assurance, and the design of models that are both expressive enough to capture operational demands and tractable enough to support rigorous analysis.

\paragraph{Accountability}

Accountability is another key requirement for real-world autonomous systems because actions and decisions must be traceable to the components, functions, or processes that originated them, so that evidence can be provided about how and why particular outcomes occurred. This is essential not only for diagnosis and system improvement, but also for assurance, post-incident investigation, and the allocation of responsibility. At the same time, accountability cannot be treated as a single, uniform notion, since the relevant audience must first be identified. In different contexts, accountability may need to be provided to regulators, stakeholders, human users, affected third parties, or even other autonomous systems, and the form in which it is presented or communicated may vary accordingly. Furthermore, accountability is closely connected to standards and legal frameworks, since the ability to explain and justify system behaviour is often necessary to demonstrate compliance, support certification, and address questions of liability. For these reasons, accountability should be treated as an explicit engineering concern in the design of real-world autonomous systems rather than as an afterthought following deployment.

\paragraph{Reliability}

Reliability in real-world autonomous systems is challenging because autonomous behaviour can make system operation difficult to predict in the face of the variability of real-world environments. This challenge is compounded by the frequent presence of black-box components, particularly machine learning modules, whose reliability may be difficult to establish with confidence. Reliability is also closely tied to resilience, since an autonomous system's ability to continue operating appropriately amid disruption cannot be easily separated from the broader system and environment in which it is embedded. Safety assurance introduces a further tension. To guarantee safe behaviour, it may be necessary to constrain autonomy, but if this abstraction is handled poorly, the result may be a system that is safe only because it is no longer meaningfully autonomous. More fundamentally, reliability depends on whether autonomous components make sound, dependable choices, linking it to notions of rationality, since systems that reason and choose more rationally are generally more trustworthy. Finally, because autonomous systems are often distributed and component-based, reliability must also be understood in compositional terms. In such systems, overall reliability is heavily influenced by the least reliable component. This marks an important difference from more monolithic systems and underscores the need for methods that can reason about reliability across heterogeneous interacting parts~\cite{Cardoso2020,Cardoso20e}.

\subsection{Future Challenges and Directions}

A central question for engineering real-world autonomous systems is to understand what can already be achieved with confidence, what remains out of reach, and what lies in the space between these two extremes. At present, the clearest success cases tend to be systems deployed in constrained environments, such as robots operating in closed factories or warehouses, where the environment, tasks, and interactions can be bounded tightly enough to support rigorous engineering and assurance. These settings have provided important evidence that reliable autonomy is possible when the operational assumptions are sufficiently controlled. However, they also make clear how far the field still is from supporting equally strong claims in open, dynamic, and socially complex environments.

One important challenge is that current engineering practice often relies on substantial simplifications of the environment. Such abstractions are necessary to make modelling, analysis, testing, and assurance tractable, but they also create uncertainty about how well the resulting evidence translates to real-world deployment. This leads directly to difficult questions about testing and validation. How many tests are enough, and what does meaningful test coverage look like for systems that may encounter changing contexts, imperfect data, novel situations, and complex stakeholder expectations? These questions are especially important when systems include machine learning components, interact closely with humans, or must operate under safety or regulatory constraints.

The discussion at the workshop highlighted the need for a clearer long-term direction for the field. We need to ask where autonomous and AI-based techniques genuinely make sense, and whether their most appropriate role may sometimes be limited to parts of a system that are not themselves safety-critical. This is not necessarily a weakness. It may instead reflect a sensible engineering boundary for the current state of the art. At the same time, if this is the current position, the field needs to articulate the major milestones that would enable broader, more trusted use of autonomy. This includes identifying where we are now, where we want to go, and which scientific, engineering, and assurance challenges must be addressed to move from one to the other.

A further direction concerns the role of formal methods within this landscape. Rather than assuming that formal methods are always the answer, we should first ask what questions they are best suited to answer in the context of autonomous systems. In some cases, formal methods may provide strong guarantees about selected components, abstractions, or decision-making layers. In other cases, they may be most valuable when combined with testing, simulation, runtime monitoring, or structured safety arguments. A key future challenge is therefore to determine how formal methods should be positioned within broader assurance strategies and what kinds of evidence they can realistically provide for real-world autonomous systems.


\section{Software Architectures for Safe Autonomous Systems}
\label{sec:architectures}

This section summarises the workshop discussion on software architectures for safe autonomous systems from three complementary perspectives: (i) practical case studies, (ii) scientific advancements, and (iii) assurance. We focus on engineering transparent decision making (and explainability), verification aligned with assurance arguments and standards, formal specification and requirements approaches, and compositional verification across heterogeneous components. In this section, we focus on the MAPE-K and BDI paradigms as representative architecture abstracts for autonomous systems. These are certainly not the only viable architecture families and other approaches include behaviour trees (e.g. \cite{tadewos2019automatic}), ROS architectures (e.g. \cite{al2024ros}), runtime assurance/safety cases (e.g. \cite{hawkins2021guidance}), state machines  (e.g. \cite{kurt2013hierarchical}) and digital twins (e.g. \cite{rosen2015importance}).

\subsection{The Relationship Between MAPE-K and BDI}
\label{sec:mk-bdi}

A recurring architectural challenge in autonomous systems is combining two forms of decision-making: adaptation to changing conditions and goal-directed deliberation about what the system should do next. MAPE-K and BDI provide two established ways of structuring these concerns. They come from different traditions (self-adaptive systems and agent-based autonomous systems, respectively) but are highly complementary for engineering safe, explainable, and resilient autonomy.

The MAPE-K loop~\cite{MAPE-K:2003} structures adaptive behaviour around five elements: \emph{Monitor}, \emph{Analyse}, \emph{Plan}, and \emph{Execute}, over shared \emph{Knowledge}. It is typically used as a conceptual reference architecture for self-adaptive systems, abstracting over concrete mechanisms for sensing, reasoning, and actuation.

Belief--Desire--Intention (BDI)~\cite{rao:95b} architectures model autonomous agents in terms of explicit mental attitudes (beliefs, goals/desires) and a deliberation/control loop over \emph{intentions} operationalised by plans (e.g., event--condition--action rules). BDI foregrounds goal management, rational choice among alternatives, and explainable action selection through explicit reasoning artefacts.

Both MAPE-K and BDI are mature and widely adopted. MAPE-K provides a clear separation of concerns for adaptation; BDI provides a principled account of intention-driven behaviour and goal change. Table~\ref{tab:mk-bdi-compare} summarises their main differences and highlights why they are useful together: MAPE-K makes adaptation structure explicit, while BDI makes the reasons for autonomous choices explicit. Combining them can yield adaptive systems with improved \emph{explainability}, \emph{ethics-by-design} (via explicit decision criteria), \emph{resilience}, and \emph{safety}. Prior work has already explored combinations of MAPE-K with control-theoretic and machine-learning elements~\cite{MAPE-CT-ML:2021}; here we highlight complementary patterns with BDI.

\begin{table*}[t]
\centering
\caption{MAPE-K vs BDI: roles and integration opportunities}
\label{tab:mk-bdi-compare}
\begin{tabularx}{\textwidth}{@{}lXX@{}}
\toprule
\textbf{Aspect} & \textbf{MAPE-K} & \textbf{BDI} \\
\midrule
Primary role & Conceptual reference model for adaptation & Concrete agent programming/execution model \\
Information flow & Explicit \emph{Monitor} stage; knowledge shared via \(K\) & Beliefs updated by perceptions; source of beliefs often implicit \\
Decision focus & Maintaining a stable/safe operating envelope; policy-level changes & Proactive goal pursuit; switching strategies by goal/plan choice \\
Granularity & Often supervisory over a managed system & Can operate at both supervisory and task levels \\
Explainability & Traceability via loop stages and artefact logs & Natural explanations via beliefs, goals, and chosen plans \\
Compositionality & Suits orchestration of heterogeneous components & Suits multi-agent coordination and goal decomposition \\
\bottomrule
\end{tabularx}
\end{table*}

We do not intend MAPE-K and BDI to be read as the only viable architectural families for safe autonomous systems. Rather, we use them here as representative abstractions that make two recurring concerns explicit: adaptation to changing conditions and goal-directed deliberation. Other architectural approaches, including behaviour trees, ROS-based architectures, runtime assurance or safety-cage patterns, state machines, blackboard architectures, and digital twins, are also widely used and may be more appropriate in particular domains. The value of focusing on MAPE-K and BDI in this section is that they provide useful conceptual handles for discussing explainability, adaptation, assurance, and compositional reasoning across heterogeneous autonomous systems.

\paragraph{Integration patterns.}
The comparison in Table~\ref{tab:mk-bdi-compare} suggests several useful integration patterns, illustrated later in the case studies:
\begin{enumerate}
    \item \textbf{BDI-in-the-Loop (A/P).} Use a BDI agent to realise the \emph{Analyse/Plan} phases of MAPE-K; the managed system executes plans while the BDI agent reasons about goal satisfaction, preferences, and trade-offs.
    \item \textbf{MAPE-K Supervises BDI.} A MAPE-K loop adapts a running BDI agent by updating its plan library, intention-selection strategy, or goals (e.g., under degradation or environmental change).
    \item \textbf{Layered Control.} BDI operates at a higher level to switch between multiple MAPE-K controllers (e.g., mission modes), while each MAPE-K controller maintains stability and safety for its mode.
    \item \textbf{BDI for Safety Shells.} A BDI agent functions as an independent safety system supervising a learning-based or reactive controller, intercepting unsafe actions and issuing compensatory behaviours.
\end{enumerate}

\subsection{Perspectives from Practical Case Studies}
\label{sec:practical-cases}

\paragraph{Pipeline Inspection (SUAVE).} The SUAVE exemplar~\cite{suaveExemplar} employs metacontrol over skills described in a TOMASyS meta-model (a concrete MAPE-K realisation). A BDI-based metacontrol implementation has been demonstrated in which belief updates reflect skill performance and environmental cues, and plan selection enacts reconfiguration decisions. This exemplifies Pattern~1 (BDI-in-the-Loop).

\paragraph{Autonomous Underwater/Drones Missions.} BDI mission scripts for BlueROV and UAVs show practical goal management with ROS integration (e.g., \texttt{mission.asl} controllers). Here, a supervisory MAPE-K loop can adjust mission goals under battery, communication, or environmental constraints (Pattern~2).

\paragraph{Nuclear Inspection.} In the nuclear inspection architecture~\cite{nuclearInspectionCaseStudy}, an intelligent safety system operates in parallel with a control stack, with independence and redundancy mandated by sector practice. A BDI safety agent supervising a MAPE-K-managed control system demonstrates Pattern~4.

\paragraph{Human--Robot Collaboration and Domestic Assistance.} Case studies on handover tasks and domestic assistants illustrate the need for traceable plan selection (BDI) and envelope maintenance (MAPE-K) to guarantee human-aware safety and predictable interaction under uncertainty.

\subsection{Perspectives on Scientific Advancements}
\label{sec:scientific-advances}

We identify four architecture-centric research directions:

\begin{enumerate}
    \item \textbf{High-level Component Contracts.} Provide machine-checkable contracts for (ROS) components and skills capturing assumptions, guarantees, and risk/uncertainty profiles (including ML components). Contracts should be consumable by both MAPE-K analysers and BDI deliberation.
    \item \textbf{Compositional Integration Patterns.} Curate and verify reusable patterns for glueing components (e.g., \emph{perception filter + certified planner + safety shell}), with proof obligations and reusable assurance claims.
    \item \textbf{Change Handling and Resilience.} Encode failure modes, degradation, and learning-triggered changes as first-class architectural concerns; expose them to both MAPE-K (for envelope maintenance) and BDI (for exceptional handling and goal revision).
    \item \textbf{Bridging the Reality Gap.} Link architectural models to implementations and runtime monitors to ensure that what is specified is what is built and what runs (cf.\ the ``as-intended vs as-implemented'' gap~\cite{coresenseD2.1}).
\end{enumerate}

\subsection{Perspectives on Assurance}
\label{sec:assurance}

Architectural decisions and patterns should be explicit artefacts in assurance arguments, enabling:
\begin{itemize}
    \item \textbf{Traceability} from hazards and safety requirements to architectural mechanisms (e.g., which layer enforces which constraint).
    \item \textbf{Compositional claims} where certified components contribute evidence to system-level goals (e.g., safety envelopes, fail-operational behaviour).
    \item \textbf{Dynamic/continuous assurance} for autonomous systems, where runtime monitors, online tests, and reconfiguration decisions update the live assurance case as the system and its environment evolve.
\end{itemize}
Relevant standards include software and system quality models and sector-specific safety norms; aligning architectural views to such standards eases certification-oriented reviews.

\subsection{Relating MAPE-K and BDI to Key Challenges}
\label{sec:mk-bdi-challenges}

\begin{description}
\item[Explainability/Transparency.] MAPE-K offers structured traceability via its loop and knowledge artefacts; BDI adds human-understandable justifications by exposing beliefs, goals, and chosen plans. Together, they support decisions that are both traceable and interpretable.
\item[Ethics/Responsibility.] BDI can encode explicit ethical constraints and preferences in goal/prioritisation policies; MAPE-K can switch operational modes to preserve ethical envelopes under context changes.
\item[Resilience/Robustness under Uncertainty.] MAPE-K maintains stability and safety margins; BDI handles exceptions and revises goals when assumptions break. Combined, they support graceful degradation and recovery.
\item[Safety.] MAPE-K implements safety envelopes and supervisory control; BDI-based safety shells intercept unsafe choices and explain interventions.
\end{description}

\subsection{Actionable Guidance}
\label{sec:actionable}

\noindent For practitioners:
\begin{enumerate}
    \item Start with a \emph{reference pattern} (e.g., planner + safety shell + MAPE-K supervisor), then add BDI for mission-level explainability and goal management.
    \item Attach \emph{contracts} to perception/ML components (assumptions, operating ranges, confidence) and surface them to both MAPE-K analyses and BDI beliefs.
    \item Instrument runtimes to \emph{log} MAPE-K artefacts and BDI deliberations; feed these into continuous assurance.
\end{enumerate}

\noindent For researchers:
\begin{enumerate}
    \item Formalise and verify the integration patterns in \S\ref{sec:mk-bdi}, with proof obligations and reusable assurance arguments.
    \item Develop tool support to co-design MAPE-K and BDI artefacts, including automated synthesis of monitors and explainers from architectural models.
\end{enumerate}

\section{Research Roadmap}


In this section, we collate and summarise the challenges discussed in the workshop, along with a roadmap for the future of Engineering Reliable Autonomous Systems.

\subsection{Challenges}

This section synthesises the main challenges that were identified across the workshop discussions and case studies. Rather than introducing new material, it distils the recurring difficulties observed across verification and validation, system engineering, and architectural design, and highlights the fundamental barriers to building reliable autonomous systems.

\subsubsection{Techniques for Verification and Validation of Autonomous Systems}

A central challenge in verification and validation is the difficulty of modelling the environments in which autonomous systems operate. Unlike traditional systems, autonomous systems are deployed in open, dynamic, and partially observable environments, where uncertainty and unpredictability are inherent. Capturing these environments in a way that is both realistic and amenable to formal reasoning remains a major obstacle.

This difficulty is compounded by the increasing use of learning-enabled components, particularly in perception and decision-making. Such components lack explicit, interpretable semantics and are highly sensitive to data distributions, making them resistant to traditional verification techniques. While progress has been made in analysing neural networks in isolation, integrating these results into system-level guarantees remains an open problem.

Another key issue is the gap between models and reality. Verification is often performed on abstract models or simulations, yet real-world deployments introduce noise, hardware constraints, and unforeseen interactions that are difficult to anticipate. This “reality gap” limits the confidence that can be placed in verification results obtained at design time.

As a consequence, there is a growing need for runtime verification and monitoring techniques that complement static analysis. However, designing monitors that are both expressive and efficient, and that can support meaningful assurance arguments, is itself a challenge. More broadly, the problem of combining evidence from multiple sources (such as formal verification, simulation, testing, and physical experiments) into coherent and convincing assurance cases remains unresolved.

Finally, the compositional nature of autonomous systems introduces additional complexity. These systems are typically constructed from heterogeneous components, including symbolic reasoning modules, machine learning models, and physical processes. Ensuring that system-level properties are preserved when integrating these components is a significant and largely unsolved challenge.

\subsubsection{Engineering Real-World Autonomous Systems}

Moving from controlled environments to real-world deployment exposes a different class of challenges. Autonomous systems are increasingly used in safety-critical domains such as healthcare, nuclear inspection, and aerospace, where failures can have severe consequences and regulatory requirements are stringent. In such contexts, engineering practices must balance innovation with rigorous assurance, often under significant uncertainty.

A major difficulty lies in system integration. Real-world autonomous systems combine multiple subsystems (such as robotic platforms, sensing pipelines, AI components, and control mechanisms) that are often developed independently. This heterogeneity increases the risk of unexpected interactions and emergent behaviours that are difficult to predict and validate.

Human--system interaction further complicates the engineering process. Many autonomous systems operate in close collaboration with humans, requiring not only functional correctness but also predictability, transparency, and trustworthiness. Designing systems that behave safely and intuitively in the presence of human users remains a substantial challenge, particularly when autonomy is only partial or when decision-making authority is shared.

Adaptation and autonomy introduce additional tension. Autonomous systems are expected to respond to changing environments and conditions, yet such adaptability can undermine predictability and complicate assurance. Ensuring that systems remain within acceptable safety and performance bounds while adapting over time is a key open problem.

Practical constraints, including scalability, real-time performance requirements, and data dependence, exacerbate these challenges. The behaviour of many systems is strongly influenced by training data, raising concerns about bias, coverage, and robustness. At the same time, constructing realistic testing and validation environments that adequately reflect operational conditions is both costly and technically demanding.

\subsubsection{Software Architectures for Safe Autonomous Systems}

At the architectural level, the challenge is to design structures that can support the desired system qualities (such as safety, explainability, and resilience) while accommodating the complexity of autonomous behaviour. Existing paradigms, such as MAPE-K and BDI, provide complementary strengths, but integrating them in a principled and verifiable manner remains an open research direction.

One of the key difficulties is enabling explainable decision-making. While symbolic approaches offer transparency, they may lack scalability, whereas learning-based approaches provide performance at the cost of interpretability. Architectures must therefore reconcile these tensions, often through hybrid or layered designs.

Another major challenge concerns the integration of heterogeneous components. Autonomous systems frequently combine symbolic reasoning, machine learning, and low-level control, each with different assumptions and guarantees. Establishing clear interfaces, contracts, and interaction patterns between these components is essential but still underdeveloped.

Architectures must also explicitly address change. Autonomous systems are subject to failures, environmental variation, and learning-driven updates, all of which require dynamic reconfiguration. Designing architectures that can handle such change while preserving safety and correctness is a significant challenge, closely linked to the need for runtime assurance.

In safety-critical domains, architectural decisions must additionally support separation of concerns, for example, through independent safety systems or monitoring layers. However, achieving the right balance between redundancy, independence, and system complexity is nontrivial.

Finally, there is the issue of traceability. Ensuring that high-level requirements, assurance arguments, and system behaviours remain aligned across design-time models and runtime execution is difficult, particularly in evolving systems.

\subsubsection{Cross-Cutting Challenges}

Across all three areas, several overarching challenges emerge. Perhaps the most prominent is the problem of assurance and certification. Existing standards and assurance frameworks are not well-suited to systems that are adaptive, learning-enabled, and deployed in open environments, creating a gap between technological capability and regulatory acceptance.

Closely related is the issue of trustworthiness. Autonomous systems must not only be safe but also predictable, explainable, and aligned with human values to gain acceptance from users, stakeholders, and society at large. This introduces ethical~\cite{Ethics:RAS:2015,Bremner19} and social dimensions that extend beyond purely technical considerations.

Another fundamental challenge is the need for interdisciplinary integration. Addressing the reliability of autonomous systems requires expertise from formal methods, artificial intelligence, robotics, human factors, and domain-specific engineering. Bridging these communities, as the workshop initiated, is essential but remains an ongoing effort.

There is also a lack of standardisation and shared benchmarks. Without common evaluation frameworks, it is difficult to compare approaches, measure progress, or transfer techniques from research to practice. This contributes to a broader gap between academic advances and industrial adoption.

Finally, autonomous systems challenge traditional notions of system lifecycle. These systems evolve, adapt to new conditions, and may change their behaviour after deployment. This necessitates a shift towards continuous or dynamic assurance, where verification and validation are ongoing processes rather than one-off activities.

\subsection{Roadmap/Pathways to Solutions }
\label{subsec:roadmap}
\begin{figure}
    \centering
    \includegraphics[width=\linewidth]{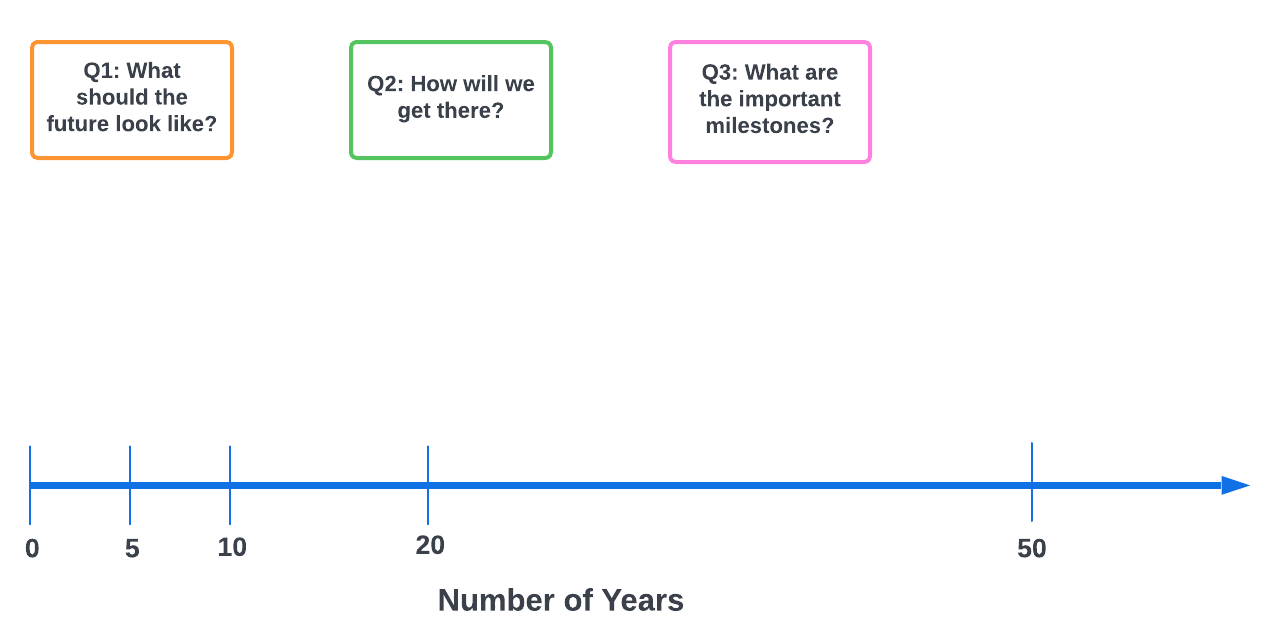}
    \caption{Template blank roadmap}
    \label{fig:template}
\end{figure}

During the workshop, participants were split into three groups and asked to fill in a blank research roadmap~(shown in Fig. \ref{fig:template}). Specifically, participants were asked the three questions:
\begin{enumerate}
    \item[\textbf{Q1}:] What should the future look like?
    \item[\textbf{Q2}:] How will we get there?
    \item[\textbf{Q3}:] What are the important milestones?
\end{enumerate}

\begin{figure}[t]
    \centering
    \includegraphics[width=\linewidth]{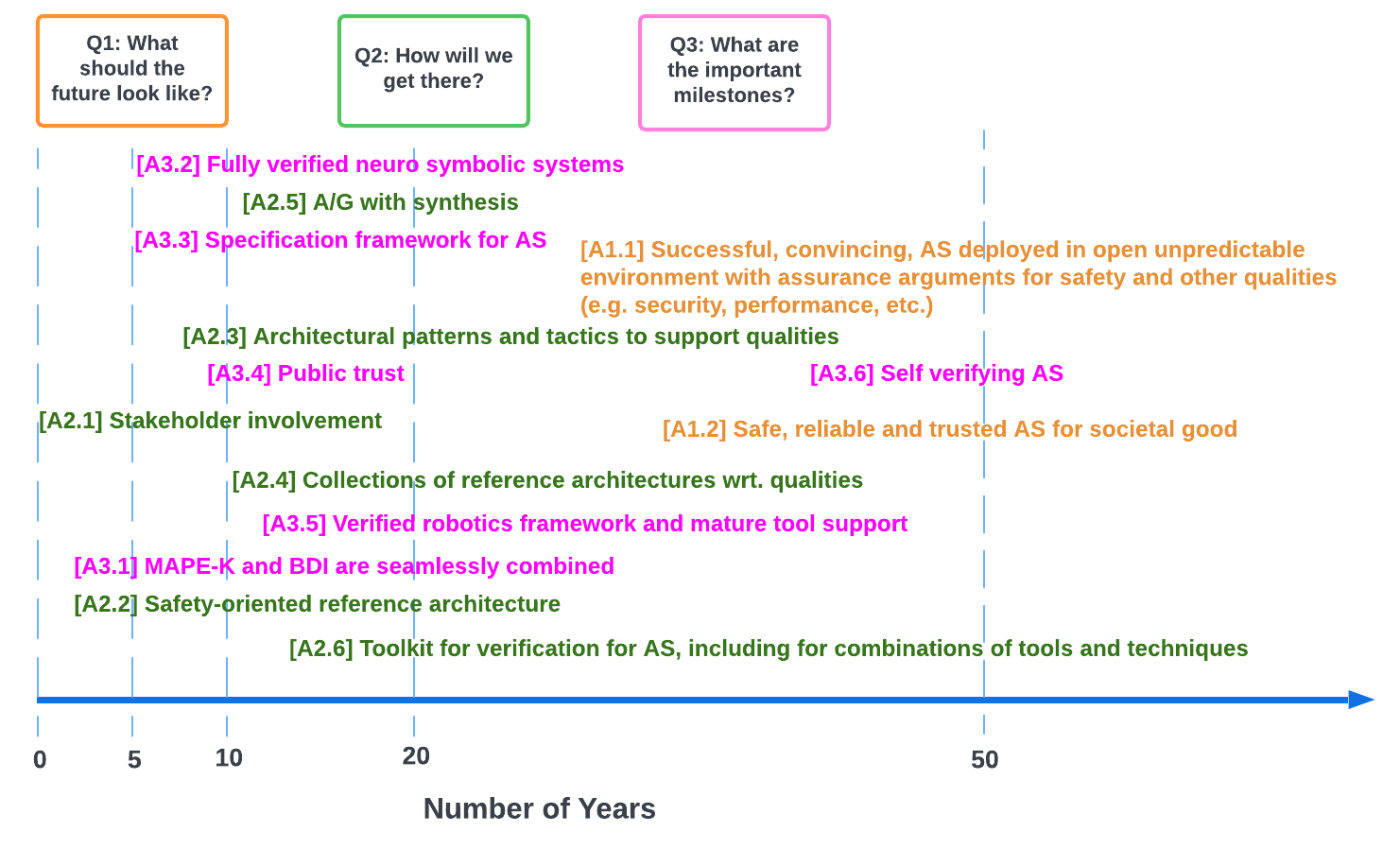}
    \caption{We consolidate the responses into a single roadmap, with colour-coding to distinguish contributions from each question. Orange denotes responses to \textbf{Q1}, green to \textbf{Q2}, and pink to \textbf{Q3}. Each response is labelled \textbf{[AX.Y]}, where \textbf{X} indicates the question and \textbf{Y} identifies the specific answer.
}
    \label{fig:mergedRoadmap}
\end{figure}

\noindent We have condensed the responses to this task as a single roadmap as shown in Fig. \ref{fig:mergedRoadmap}. These responses are colour-coded in reference to the individual questions. Specifically, orange points are in response to \textbf{Q1}, green points refer to \textbf{Q2}, and pink points correspond to \textbf{Q3}. We reflect on the responses per question in the following subsections. We also label each of the answers as \textbf{[AX.Y]} where \textbf{X} corresponds to the question number and \textbf{Y} identifies a unique answer.

\subsubsection{Q1: What Should The Future Look Like?}
\label{subsubsec:q1}
The workshop participants overwhelmingly recognised the importance of interdisciplinary research and collaboration:~not only to develop tools and techniques for verifying, validating, and demonstrating reliable autonomous systems, but also to devise scalable and practical development processes. Such efforts are already underway worldwide, but the pace must be accelerated if we are to improve the reliability of rapidly evolving autonomous systems.

One area of particular focus in the coming years is the verification of vision-based perception systems. Since many autonomous systems that interact with the real, physical world rely on object detection and recognition (including autonomous vehicles and domestic robotic assistants), it is vital that we can verify the accuracy of perception systems. This will likely require heterogeneous techniques for verification and validation, including formal methods, and simulation and testing-based approaches, to provide acceptable levels of assurance that perception systems function correctly \cite{Farrell2018,WWADEMP19}. The current state-of-the-art combines various logics (e.g. temporal and spatial \cite{hekmatnejad2021phdthesis,hekmatnejad2024ijrr}) and proposes new logics (e.g. for bounding boxes \cite{tanaka2025access}) to attempt to bridge this gap.

Of course, even if we devise mature and sophisticated verification techniques, in safety-critical domains (e.g., nuclear and aerospace), we must be able to convince the relevant regulatory authorities that we have ensured the reliability of a given autonomous system. For inhospitable environments, such as nuclear ones \cite{luckcuck_matt_2021_5012322}, human-is-not-always-in-the-loop and high-fidelity simulations will inevitably be as close as we can conceivably get to reality before deployment. Participants remarked that it will take significant time to convince regulators and for regulatory frameworks to catch up to the advances in autonomous systems (although the benefits far outweigh the alternatives in certain areas). 

That said, several international standards have been developed to provide initial guidance in this area, including ISO 26262 \cite{ISO26262} for autonomous vehicles and IEEE 7009 \cite{IEEE7009} for fail-safe design of autonomous systems. The development of standards related to autonomous systems (including the development processes used) is fertile ground at present, with several initiatives forthcoming (e.g., IEEE P7009.1 \cite{IEEE7009-1}).

Although our discussions focused on safety-related aspects, it is clear that autonomous system reliability encompasses several other qualities, including security, ethics, and performance. This list is likely to evolve as the diverse benefits and concerns of autonomous systems reveal themselves across application domains. As a result, it will be necessary to develop standards, certification and assurance arguments for all of these important qualities.

We foresee a future in which autonomous systems are used for the societal good. A world where verification and assurance are first-class citizens in the development of autonomous systems that can work seamlessly and reliably both alongside and in the absence of humans to complete useful, critical and potentially dangerous tasks to support humans in advancing scientific discovery, quality of life, healthcare, emergency services, etc.

\subsubsection{Q2: How Will We Get There?}
\label{subsubsec:q2}


The question of how we get to a world where autonomous systems are developed to be reliable and deployed in widely varying contexts is not an easy one to answer. In fact, there are undoubtedly several (potentially complementary) routes to achieve this. 

Central to our discussions was the need to involve a range of stakeholders who can help ensure that cutting-edge research in this area has a practical and realisable impact. Notably, many of these stakeholders will not be experts in autonomous systems, many may not even fully realise the benefits that autonomy can bring to their organisations, application areas and products. Collaborations with these diverse stakeholders will create closer links between academia and industry and richer sets of scalable tools and techniques that can cope with the complexity associated with autonomous systems. 

At a technical level, this work will deliver: (i) reference architectures for safety, privacy, ethical behaviour, and related concerns; (ii) methods for synthesising autonomous system implementations from verified specifications and formally defined requirements; and (iii) a practical toolkit for linking heterogeneous verification artefacts, enabling a holistic approach to the verification and assurance of autonomous systems. These advances will be complemented by targeted improvements to the underlying verification and assurance tools themselves. 

The general public is a very important stakeholder, and societal acceptance will be achieved through educational programs, science communication, successful small-scale pilot programs, and ethical arguments (e.g., it is better to place an autonomous robot in a nuclear reactor than a human). We are already seeing societal acceptance of autonomous driving begin in certain cities \cite{hewitt2019assessing}, but these remain in relatively controlled experiments, and accidents still happen\footnote{Accessed: 17 June 2026 \url{https://edition.cnn.com/us/waymo-robotaxis-safety-invs}}
\footnote{Accessed: 17 June 2026 \url{https://www.reuters.com/business/autos-transportation/gm-self-driving-unit-cruise-pay-15-million-fine-over-crash-disclosure-2024-09-30/}} 
\footnote{Accessed: 17 June 2026 \url{https://edition.cnn.com/2025/10/10/business/tesla-investigation-self-driving-intl}} \cite{wang2020safety}. 

Since absolute safety can never be guaranteed for any system that interacts with an unpredictable environment, it will also be necessary to put in place appropriate legal frameworks that establish responsibility, liability, and accountability \cite{burton2020mind,pagallo2017automation}. Such frameworks will help to reassure the public that autonomous systems should obey reasonable rules and laws. That said, all of these take political and institutional will.






\subsubsection{Q3: What Are The Important Milestones?}
\label{subsubsec:milestons}

There are several important milestones along the way, some of which are identified in Fig. \ref{fig:mergedRoadmap}. In the near term, the seamless combination of the MAPE-K and BDI paradigms will integrate control loops with intentional agent reasoning. This will support the development of systems that both dynamically respond to changes and exhibit goal-driven behaviour within a coherent architecture. This combination is already discussed in Section \ref{sec:architectures}. 

Beyond this, fully-verified neuro-symbolic systems will ensure that high-level rational reasoning components and learning-enabled sub-symbolic forms of AI will be mathematically proven to behave as expected \cite{lu2024surveying,belle2023neuro}. This will enable systems that are both highly capable and mathematically assured to behave correctly, safely, securely, etc. This will be complemented by the development of specification frameworks for autonomous systems, providing a holistic approach to defining the formal requirements they must meet. This will support systematic design, verification and validation by providing a shared language between developers, regulators and verification tools/techniques, likely building on pre-existing approaches to assume-guarantee reasoning and compositional verification \cite{elkader2015automated,blundell2005assume,Chilton13,Cardoso20e,Cobleigh2003,cofer2012compositional,Spellini2019,10.1145/3092282.3092290,pinto2023leveraging}. Such a specification framework will enable an associated verified robotics framework and mature tool support. This will provide formally verified models and components that guarantee correct operation across the sensing, planning, and actuation layers. For this, mature tool support is vital to ensure integration into real-world development pipelines through well-tested software ecosystems. 

These milestones will each help engender public trust in autonomous systems that are expected to consistently demonstrate safety, transparency, and ethical behaviour in real-world scenarios. This will help pave the way for the development of, and trust in, self-verifying autonomous systems that continuously assess and validate their own behaviour during operation against formal specifications and safety constraints. This capability supports sophisticated runtime assurance, allowing the system to detect deviations and adapt or fail safely without external intervention.

Whilst our roadmap presented in this section focuses on positive steps that we must take to ensure the reliability of autonomous systems. There are several non-ignorable concerns about the use of autonomous systems. The first, in particular, concerns the use of AI systems like Large Language Models, which consume vast amounts of energy and thus have implications for environmental sustainability \cite{singh2025survey}. Overuse of such AI systems may significantly exacerbate the climate crisis. Another concern is that humans in many roles will become obsolete due to the prevalence of AI. Although this is a positive feature in situations where humans are removed from harm, negative consequences include mass unemployment, increased poverty, antisocial behaviour, and unrest \cite{pettigrew2018potential,bissell2020autonomous}.

\subsubsection{Summary}

Our current roadmap considers what the future should look like, how to get there, and which milestones matter, but it would benefit from more clearly defined deliverables. We identify several potential deliverables in Table \ref{tab:timehorizons} that are related to the milestones described in Fig. \ref{fig:mergedRoadmap}. This is not a complete list, rather we intend it as a starting point for future research  projects. In the near-term we advocate for (1) curated benchmark case studies that can be used to drive and evaluate development, (2) standardised terminology that is accessible to computer scientists, engineers and roboticists alike, (3) standardised scoring rubrics that can be used to evaluate aspects including (but not limited to) expressivity, efficiency, verifiability, etc. and (4) evidence templates for autonomous systems assurance that can be adopted and used effectively across diverse application domains. In the medium-term we encourage the development of (1) component-contract specification languages that can capture the peculiarities and uncertainty that is unique to autonomous systems, (2) techniques for synthesising monitors from formal specifications that can capture evolving/dynamic autonomous systems at runtime, (3) reusable assurance case patterns that build upon the near-term evidence templates, and (4) tool interoperability including middlewares, specification paradigms, verification tools, etc. In the long-term we aspire to build (1) self-monitoring and self-verifying autonomous systems, (2) achieve broad regulatory acceptance in compliance with (what will become) mature industrial standards for autonomous systems, and (3) fully certified (across multiple domains) neuro-symbolic autonomous functionality.

\begin{table}[t]
    \centering
    \begin{tabular}{p{4.6cm} p{4.7cm} p{4.6cm}}
    \hline
        \textbf{Near-Term} & \textbf{Medium-Term}  & \textbf{Long-Term }\\
        \textbf{(0--10 years)} & \textbf{(10--20 years)}  & \textbf{(20+ years)}\\\hline
        \rowcolor{lightgray!60} Curated Benchmark Case Studies & Component-Contract Specification Languages & Self-Monitoring and Self-Verifying AS\\
         Standardised Terminology & Monitor Synthesis & Broad Regulatory Acceptance \\
        \rowcolor{lightgray!60} Standardised Scoring Rubrics & Reusable Assurance Case Patterns & Certified Neuro-Symbolic Autonomy \\
         Evidence Templates & Tool Interoperability & \\
    \end{tabular}
    \caption{Deliverables: Near-, Medium- and Long-Term.}
    \label{tab:timehorizons}
\end{table}

\section{Conclusion}
\label{sec:conclude}

Autonomous systems are increasingly being considered for deployment in domains where failures may have severe consequences, including healthcare, nuclear inspection, space operations, underwater missions, industrial automation, domestic assistance, and emergency response. The promise of these systems is significant: they can extend human capability, remove people from hazardous environments, and support complex decision-making in settings where conventional automation is insufficient. However, the same features that make autonomous systems valuable also make them difficult to engineer, verify, validate, and assure. Their behaviour depends on uncertain environments, heterogeneous software and hardware components, learning-enabled perception and decision-making, human interaction, and, in many cases, adaptation after deployment.

This workshop report has summarised the outcomes of the ERAS workshop, which brought together the FMAS and AREA communities to identify common challenges and future directions for engineering reliable autonomous systems. A central message from the workshop is that reliability cannot be treated as a single property or as a late-stage verification activity. Instead, it must be addressed throughout the lifecycle of an autonomous system, from requirements elicitation and architecture design through implementation, verification, deployment, monitoring, and evolution. Safety remains a primary concern, but reliability also depends on related qualities such as robustness, resilience, transparency, explainability, accountability, security, privacy, sustainability, SLEEC-awareness and trustworthiness.

The case studies discussed in this report show that progress is already being made across a wide range of application domains. They demonstrate the value of formal modelling, runtime verification, simulation, physical experimentation, safety cases, requirements engineering, and architectural separation of concerns. At the same time, they reveal persistent gaps between research prototypes and dependable real-world deployment. In particular, the ``reality gap'' between models, simulations, and physical systems remains a major obstacle. So too does the challenge of integrating evidence from diverse verification and validation techniques into assurance arguments that are convincing to engineers, regulators, users, and the wider public.

A recurring theme across the workshop was the need to combine complementary approaches. Formal methods provide rigour and precise reasoning, but must be made scalable and usable in realistic engineering processes. Agent-based and cognitive architectures provide explicit representations of goals, beliefs, plans, and decisions, but must be integrated with lower-level control, perception, and runtime assurance mechanisms. Learning-enabled components provide powerful capabilities, especially for perception and adaptation, but require new forms of specification, verification, monitoring, and explanation. Architectures such as MAPE-K and BDI offer useful foundations, and their principled integration points towards systems that are both adaptive and explainable, both responsive and goal-directed, and both capable and auditable.

The research roadmap developed during the workshop identifies several pathways towards reliable autonomous systems. These include the development of shared benchmarks and realistic case studies, stronger links between architecture and assurance, verified and reusable component contracts, better techniques for reasoning about uncertainty and change, and toolchains that connect requirements, models, code, monitors, and assurance cases. The roadmap also highlights the importance of interdisciplinary collaboration. Formal methods, robotics, artificial intelligence, software engineering, human factors, regulation, or domain expertise in isolation cannot achieve reliable autonomy. It requires sustained interaction between these communities, together with engagement from industry, regulators, policymakers, and the public.

The workshop also made clear that societal acceptance will depend not only on technical correctness, but also on justified confidence. Autonomous systems must be demonstrably safe, understandable to relevant stakeholders, aligned with legal and ethical expectations, and subject to appropriate accountability mechanisms. This is especially important as autonomy moves into domains where the human may not always be in the loop, where systems may need to adapt at runtime, and where assurance must continue after deployment.

In conclusion, engineering reliable autonomous systems is both a technical and societal challenge. The field has reached a point where isolated advances are no longer sufficient:~progress depends on integrated methods, shared artefacts, realistic exemplars, and sustained collaboration across communities. The ERAS workshop represents an important step towards this goal by bringing together researchers and practitioners who approach autonomy from different but complementary perspectives. The challenge now is to turn this shared understanding into practical methods, mature tools, credible assurance processes, and deployable systems that can realise the benefits of autonomy while maintaining the confidence of those who build, regulate, use, and are affected by them.


\clearpage
\bibliographystyle{plainurl}
\bibliography{ref}

\end{document}